\documentclass[journal]{IEEEtran}
\usepackage{algorithm}
\pdfoutput=1

\usepackage[algo2e,lined]{algorithm2e}
\usepackage{amsmath}
\usepackage{amssymb} 
\usepackage[colorlinks,bookmarksopen,bookmarksnumbered,citecolor=red,urlcolor=red]{hyperref}
\usepackage{blindtext}
\usepackage{bm}
\usepackage{booktabs}
\usepackage[sort&compress,numbers]{natbib}
\usepackage{multirow}
\usepackage{pseudocode}
\usepackage{comment}
\usepackage{graphicx}
\usepackage{xcolor}
\usepackage{subfigure}
 
\newcommand\blfootnote[1]{%
  \begingroup
  \renewcommand\thefootnote{}\footnote{#1}%
  \addtocounter{footnote}{-1}%
  \endgroup
}

\newcommand{\reffig}[1]{Fig.~\ref{#1}}
\newcommand{\refeq}[1]{Eq.~(\ref{#1})}
\newcommand{\reftab}[1]{Table~\ref{#1}}
\newcommand{\refsec}[1]{Section~\ref{#1}}

\newcommand{\refalg}[1]{Alg.~\ref{#1}}

\def\wrt{w.r.t.~}

\newcommand{\myvec}[1]{\boldsymbol{#1}}
\newcommand{\myvecf}[2]{_{#2}{\boldsymbol{#1}}}
\newcommand{\myvecff}[3]{_{#1}{\boldsymbol{#2}}_{#3}}

\newcommand{\myquat}[2]{\boldsymbol{q}_{{#1}{#2}}}
\newcommand{\mytrafo}[2]{\boldsymbol{T}_{{#1}{#2}}}
\newcommand{\myrotmat}[2]{\boldsymbol{R}_{{#1}{#2}}}
\newcommand{\myframe}[1]{\mathcal{F}_{#1}}

\newcommand{\norm}[1]{\left\lVert#1\right\rVert}

\newcommand{\calibF}{\myvec{f}}
\newcommand{\calibC}{\myvec{c}}
\newcommand{\calibW}{w}
\newcommand{\calibExtT}{\myvecff{C}{p}{CI}}
\newcommand{\calibExtR}{\myquat{C}{I}}

\newcommand{\calibMa}{\myvec{m}_{a}}
\newcommand{\calibMg}{\myvec{m}_{g}}
\newcommand{\calibSa}{\myvec{s}_{a}}
\newcommand{\calibSg}{\myvec{s}_{g}}
\newcommand{\calibExtAG}{\myquat{A}{I}}

\newcommand{\spaceReal}{\mathbf{R}}
\newcommand{\spaceRealX}[1]{\spaceReal^{#1}}
\newcommand{\spaceRot}{SO(3)}
\newcommand{\spaceSE}{SE(3)}

\newcommand{\threeDOF}{3-DoF~}
\newcommand{\sixDOF}{6-DoF~}

\definecolor{todo-red}{RGB}{200,12,12}
\definecolor{green4}{RGB}{0,128,0}

\newcommand{\q}[2]{\boldsymbol{q}_{{#1}{#2}}}

\newcommand{\Cq}[2]{\boldsymbol{C}(\q)}

\usepackage[printonlyused,withpage,nolist,nohyperlinks]{acronym}
\begin{acronym}
\acro{2D}{two-dimensional}
\acro{3D}{three-dimensional}
\acro{DoF}{degree of freedom}
\acro{IMU}{inertial measurement unit}
\acro{EKF}{extended Kalman filter}
\acro{SLAM}{simultaneous localization and mapping}
\acro{VI}{visual-inertial}
\acro{VIO}{visual-inertial odometry}
\acro{MEMS}{Micro Electro-Mechanical Systems}
\acro{RMSE}{root-mean-square error}
\acro{MAV}{Micro Aerial Vehicle}
\acro{UAV}{Unmanned Aerial Vehicle}
\acro{ML}{maximum-likelihood}
\acro{SfM}{structure from motion}
\end{acronym}

\newcommand\BibTeX{{\rmfamily B\kern-.05em \textsc{i\kern-.025em b}\kern-.08em
T\kern-.1667em\lower.7ex\hbox{E}\kern-.125emX}}

\begin{document}
\title{Observability-aware Self-Calibration of Visual and Inertial Sensors for Ego-Motion Estimation}

\author{Thomas~Schneider, Mingyang~Li, Cesar~Cadena, Juan~Nieto, and Roland~Siegwart
\thanks{T. Schneider, C. Cadena, J. Nieto and R. Siegwart are with the Autonomous Systems Lab, ETH Z\"urich, Z\"urich, CH-8092, Switzerland (e-mail: \{schneith,cesarc,nietoj,rsiegwart\}@ethz.ch).}
\thanks{M. Li is with the Alibaba Group, Hangzhou, China (e-mail: mingyangli009@gmail.com).}
}

\markboth{}%
{SCHNEIDER et al.: Observability-aware Self-Calibration of Visual and Inertial Sensors for Ego-Motion Estimation}

\maketitle

\begin{abstract}
External effects such as shocks and temperature variations affect the calibration of visual-inertial sensor systems and thus they cannot fully rely on factory calibrations.
Re-calibrations performed on short user-collected datasets might yield poor performance since the observability of certain parameters is highly dependent on the motion.
Additionally, on resource-constrained systems (e.g mobile phones), full-batch approaches over longer sessions quickly become prohibitively expensive.

In this paper, we approach the self-calibration problem by introducing information theoretic metrics to assess the information content of trajectory segments, thus allowing to select the most informative parts from a dataset for calibration purposes.
With this approach, we are able to build compact calibration datasets either: (a) by selecting segments from a long session with limited exciting motion or (b) from multiple short sessions where a single sessions does not necessarily excite all modes sufficiently.
Real-world experiments in four different environments show that the proposed method achieves comparable performance to a batch calibration approach, yet, at a constant computational complexity which is independent of the duration of the session.
\end{abstract}


\blfootnote{An earlier \href{https://ieeexplore.ieee.org/document/7989766}{version} of this paper was presented at the \textit{2017 IEEE International Conference on Robotics and Automation (ICRA)} and was published in its Proceedings.}

\section{Introduction}
\label{sec:introduction}
\IEEEPARstart{I}{n} this work, we present a sensor self-calibration method for visual-inertial ego-motion estimation frameworks i.e. systems that fuse visual information from one or multiple cameras with an \ac{IMU} to track the pose (position and orientation) of the sensors over time.
Over the last years, visual-inertial tracking has become an increasingly popular method and is being deployed into a big variety of products including AR/VR headsets, mobile devices, and robotic platforms.
Large-scale projects, such as Microsoft’s HoloLens, make these complex systems available as part of mass-consumer devices operated by non-experts over the entire life-span of the product.
This transition from the traditional lab environment to the consumer market poses new technical challenges to keep the calibration of the sensors up-to-date.

\begin{figure}[ht]
	\centering
	\includegraphics[width=0.5\textwidth]{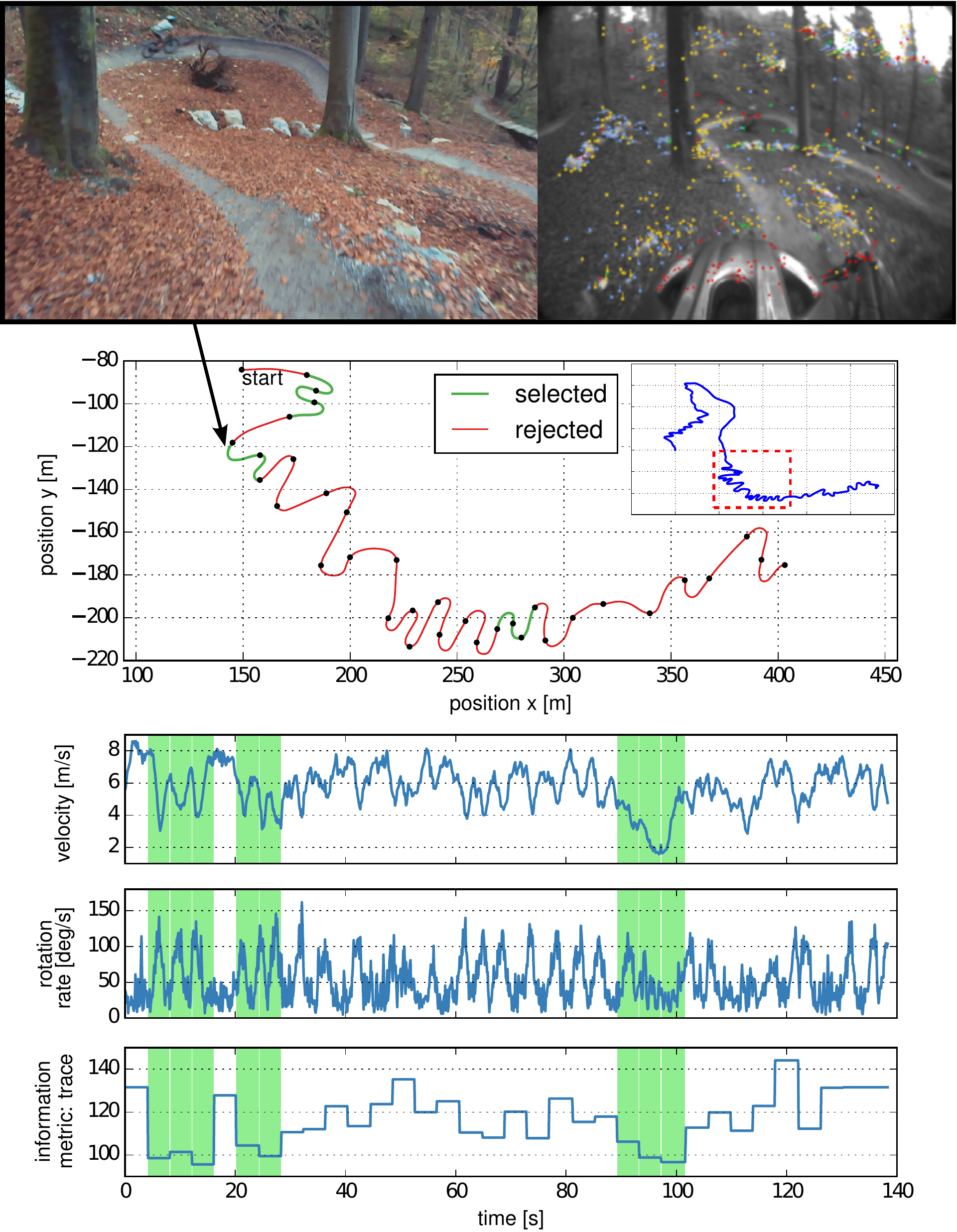}
	\caption{
	    Dataset recorded while riding down Mount Uetliberg on a mountain-bike with a Tango Tablet strapped to the rider's head.
	    This trajectory is a good example of the varying amount of information within different segments of a visual-inertial dataset.
	    Our method identifies the most informative segments in a background process alongside an existing visual-inertial motion estimation framework.
	    Consequently, we sparsify the dataset to ensure an efficient calibration of the camera and \ac{IMU} model parameters.
	    The illustration highlights the $8$ most informative segments which are sufficient for a reliable calibration.\textsuperscript{1}
	    }
	\label{fig:teaser}
\end{figure}

\blfootnote{\textsuperscript{1}Figures in this paper are best viewed in color.}

Traditionally, visual-inertial sensors are calibrated in a laborious manual process by an expert often using specialized tools and external markers such as checkerboard patterns (e.g. \cite{rehder2016extending}).
Aside from a lack of equipment, the lack of knowledge on how to properly excite all modes usually renders these methods infeasible for consumers as specific motion is required to obtain a consistent calibration.
However, it can be used at the factory to provide an initial calibration for the device.
Due to varying conditions (e.g. temperature, shocks, etc.) such calibrations degrade over time and periodic re-calibrations become necessary.
A straightforward approach to this problem would be to run a calibration over a long dataset, hoping it is rich enough to excite all modes of the system.
Yet, the large computational requirement of such a batch method might render this approach infeasible on constrained platforms without careful data selection.

This work exploits that information is usually not distributed uniformly along the trajectory of most visual-inertial datasets, as illustrated in \reffig{fig:teaser} for a mountain-bike dataset.
trajectory segments with higher excitation provide more information for sensor calibration whereas segments with weak excitation can lead to a non-consistent or even wrong calibration.
Consequently, we propose a calibration architecture that evaluates the information content of trajectory segments in a background process alongside an existing visual-inertial estimation framework.
A database maintains the most informative segments that have been observed either in a single-session or over multiple sessions to accumulate relevant calibration data over time.
Subsequently, the collected segments are used to update the calibration parameters using a segment-based calibration formulation.

By only including the most informative portion of the trajectory, we are able to reduce the size of the calibration dataset considerably.
Further, we can collect exciting motion in a background process assuming such motion occurs eventually and thus take the burden from the users to perform them consciously (which might be hard for non-experts).
With this approach we can automate the traditional tedious calibration task and perform a re-calibration without any user intervention e.g. while playing an AR/VR video game or while navigating a car through the city.
Additionally, our method facilitates the use of more advanced sensor models (e.g. \ac{IMU} intrinsics) with potentially weakly observable modes that require specific motion for a consistent calibration.

This article is an extension of our previous work \cite{schneider2017visual} where we presented the following:
\begin{itemize}
	\item an efficient information-theoretic metric to identify informative segments for calibration,
	\item a segment-based self-calibration method for the intrinsic and extrinsic parameters of a visual-inertial system, and
	\item evaluations of the calibration parameter repeatability showing comparable performance to a batch approach.
\end{itemize}
In this work, we extend with the following contributions:
\begin{itemize}
    \item a comprehensive review of the state-of-the-art on visual and inertial sensor calibration,
    \item a study of three different metrics for the selection of informative segments,
    \item an evaluation of the motion estimation accuracy on motion-capture ground-truth, and
    \item a comparison against an \ac{EKF} approach that jointly estimates motion and calibration parameters.
\end{itemize}

\section{Literature Review}
\label{sec:related_work}
Over the past two decades, visual-inertial state estimation has been studied extensively by the research community and many methods and frameworks have been presented.
For example, the work of \citet{leutenegger2015keyframe} fuses the information of both sensor modalities in a fixed-lag-smoother estimation framework and demonstrates metric pose tracking with an accuracy in the sub-percent range of distance traveled.
Many applications on resource-constrained platforms, such as mobile phones, however, use filtering-based approaches which offer pose tracking with similar accuracy at a lower computational cost.
An early method of this form is the one from \citet{mourikis2007multi}, and more recently also from \citet{bloesch2015robust}, that directly minimizes a photometric error on image patches instead of a geometric re-projection error on point-features.
Newer frameworks e.g. from \citet{qin2017vins} or \citet{schneider2018maplab} also incorporate online localization/loop-closures to further reduce the drift or in certain cases even eliminate it completely.

%
%
All these methods require an accurate and up-to-date calibration of all sensor models to achieve good estimation performance.
For this reason, a multitude of methods have been developed to calibrate models for the camera, \ac{IMU} and relative pose between the two sensors.
An overview of early methods that calibrate each model independently can be found in \cite{hartley2003multiple, alves2003camera, lobo2007relative}.
In the remaining of this section we, first, provide an overview of the state of the art in self-calibration of visual-inertial sensor systems and, second, discuss the most relevant \emph{observability-aware} calibration approaches.
And finally, we review methods that perform information-theoretic data selection for calibration purposes; which are most related to our approach.

\subsection{Marker-based Calibration}
%
%
The work on self-calibration of visual and inertial sensors is still limited and therefore, we first discuss approaches that rely on external markers such as checkerboard patterns.
An approach based on an \ac{EKF} is presented in \cite{mirzaei2008kalman} that uses a checkerboard pattern as a reference to jointly estimate the relative pose between an \ac{IMU} and a camera with the pose, velocity, and biases.
\citet{zachariah2010joint} additionally estimate the scale error and misalignment of the inertial axis using a sigma-point Kalman filter.

A parametric method is proposed in \cite{furgale2013unified} describing a batch estimator in continuous-time that represents the pose and bias trajectories using B-splines. 
\citet{krebs2012generic} extends this work by compensating additional sensing errors in the \ac{IMU} model; namely measurement scale, axis misalignment, cross-axis sensitivity, the effect of linear accelerations on gyroscope measurements and the orientation between the gyroscope and the accelerometer.
A similar model is calibrated by \citet{nikolic2016non} where they make use of a non-parametric batch formulation and thus avoid the selection of a basis function for the pose and bias trajectories which might depend on the dynamics of the motion (e.g. over the knot density).
The non-parametric and parametric formulation are compared in real-world experiments with the conclusion that the accuracy and precision of both methods are similar \cite{nikolic2016non}.

\subsection{Marker-less Calibration}
%
%
In contrast to target-based, self-calibration methods solely rely on natural features to calibrate the sensor models without the need for external markers such as checkerboards.
Early work of this from was presented by \citet{kelly2011visual} and uses an unscented Kalman filter to jointly estimate pose, bias, velocity, \ac{IMU}-to-camera relative pose and also the local scene structure.
Their real-world experiments demonstrate that the relative pose between a camera and an \ac{IMU} can be accurately estimated with similar quality to target-based methods.
The work of \citet{patron2015spline} additionally calibrates the camera intrinsics and uses a continuous-time formulation with a B-splines parameterization.
\citet{li2014high} go one step further and also include the following calibration parameters into the (non-parametric) \ac{EKF}-based estimator: time offset between camera and \ac{IMU}, scale errors and axis misalignment of all inertial axis, linear acceleration effect on the gyroscope measurements (g-sensitivity), camera intrinsics including lens distortion and the rolling-shutter line-delay.
A simulation study and real-world experiments indicate that all these quantities can indeed be estimated online solely-based on natural features~\cite{li2014high}.

\subsection{Observability of Model Parameters}
%
%
All of the discussed calibration methods so far, both target-based and self-calibration methods, rely on sufficient excitation of all sensor models to yield an accurate calibration.
\citet{mirzaei2008kalman} formally prove that the \ac{IMU}-to-camera extrinsics are observable in a target-based calibration setting where the observability only depends on sufficient rotational motion.
The analysis of \citet{kelly2011visual} shows that the \ac{IMU}-to-camera extrinsics remains observable also for a self-calibration formulation.
Further, \citet{li2014online} derive the necessary condition for the identifiability of a constant time offset between the \ac{IMU} and camera measurements.

So far, no observability analysis has been performed for the full joint self-calibration problem that includes the intrinsics of the \ac{IMU} and camera and also the relative pose between the two sensors.
Our experience, however, indicates that `rich' exciting motion is required to render all parameters observable and usually such calibration datasets are collected by expert intuition.
Often, this knowledge is missing when \ac{SLAM} systems are deployed to consumer-market products.
For this reason, the (re-)calibration dataset collection process must be automated for true life-long autonomy.

\subsection{Active Observability-aware Calibration}
Active calibration methods automate the dataset collection by planning and executing trajectories which ensure the observability of the calibration parameters wrt. a specified metric.
An early work in this direction for target-based camera calibration is \cite{richardson2013aprilcal}.
They present an interactive method that suggests the next view of the target that should be captured such that the quality of the model improves incrementally.

Another active calibration method is presented by  \citet{bahnemann2017sampling} to plan informative trajectories using a sampling-based planner to calibrate \ac{MAV} models.
The informativeness of a candidate trajectory segment within the planner is approximated by the determinant of the covariance of the calibration parameters which is propagated using an \ac{EKF}.
In a similar setting, \citet{hausman2017observability} plan informative trajectories to calibrate the model of an \ac{UAV} using the local observability Gramian as an information measure.
An extension to this work is presented by \citet{preiss2017trajectory} where they additionally consider free-space information and dynamic constraints of the vehicle within the planner.
The condition number of the \textit{Expanded Empirical Local Observability Gramian} (E\textsuperscript{2}LOG) is proposed as an information metric.
The columns of E\textsuperscript{2}LOG are scaled using empirical data to balance the contribution of multiple states.
A simulation study shows that the method outperforms random motion and also the well-known heuristic, such as the figure-8 or star motion pattern.
Further, the study indicates that trajectories minimizing the E\textsuperscript{2}LOG perform slightly better compared to the minimization of the trace of the covariance matrix but in general yield comparable performance.

\subsection{Passive Observability-aware Calibration -- Calibration on Informative Segments}
In contrast to the class of active calibration methods, passive methods cannot influence the motion and instead identify and collect informative trajectory segments to build a complete calibration dataset over time.
The framework of \citet{maye2013self} selects a set of the most informative segments using an information gain measure to consequently perform a calibration on the selected data.
A truncated-QR solver is used to limit updates to the observable subspace.
The generality of this method makes it suitable for a wide range of problems.
Unfortunately, the expensive information metric and optimization algorithm prevent its use on resource-constrained platforms.
Similarly, \citet{keivan2014constant} maintain a database of the most informative images to calibrate the intrinsic parameters of a camera but use a more efficient entropy-based information metric for the selection.
\citet{nobre2016multi} extend the same framework to calibrate multiple sensors and more recently \citet{nobre2017drift} also include the relative pose between an \ac{IMU} and a camera.

In our work, we take a similar approach to \cite{maye2013self,keivan2014constant} but also consider inertial measurements and consequently collect informative segments instead of images.
In contrast to the general method of \cite{maye2013self}, we use an approximation for the visual-inertial use-case and neglect any cross-terms between segments when evaluating their information content.
This approximation increases the efficiency at the cost that no loop-closure constraints can be considered.
Compared to \cite{nobre2017drift}, we assume the calibration parameters to be constant over a single session but additionally calibrate the intrinsic parameters of the \ac{IMU} using a model similar to \cite{krebs2012generic,li2013high}.

\section{Visual and Inertial System}
\label{sec:vi_system}
The visual-inertial sensor system considered in this work consists of a global-shutter camera and an \ac{IMU}.
For better readability, the formulation is presented only for a single camera, however, the method has been tested for multiple cameras as well.
All sensors are assumed to be rigidly attached to the sensor system.
The \ac{IMU} itself consists of a 3-axis accelerometer and a 3-axis gyroscope.
In this work, we assume an accurate temporal synchronization of the \ac{IMU} and camera measurements and exclude the estimation of the clock offset and skew.
However, online estimation of these clock parameters is feasible as shown in \cite{li2014online}.

\begin{figure}[t]
\centering
\includegraphics[width=0.38\textwidth]{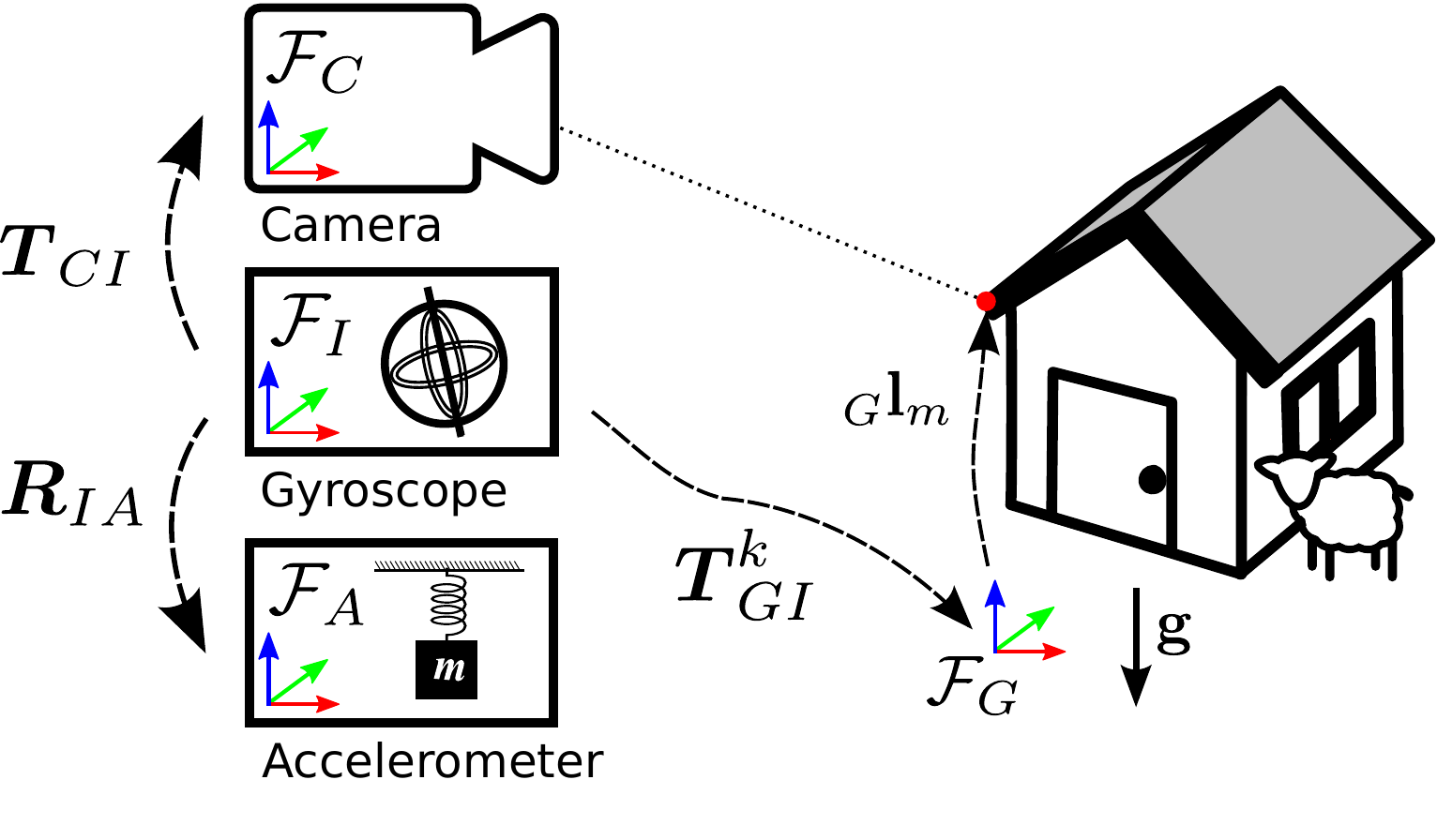}
\caption{
Coordinate frames of the visual-inertial sensor system:
The camera, \threeDOF gyroscope and \threeDOF accelerometer are all rigidly attached to the sensor system.
The frame $\myframe{C}$ denotes the frame of the camera where $\myvecff{C}{e}{z}$ points along the optical axis, $\myvecff{C}{e}{x}$ left-to-right and $\myvecff{C}{e}{y}$ top-down as seen from the image plane.
The \sixDOF transformation matrix $\mytrafo{C}{I}$ (extrinsic calibration) relates the \ac{IMU} $\myframe{I}$ (which is defined to coincide with the frame of the gyroscope) to the frame of the camera $\myframe{C}$.
Since the translation $\myvecff{I}{p}{IA}$ between the gyroscope and the accelerometer is typically close to zero for single-chip \ac{MEMS} sensors, we only rotate the accelerometer frame $\myframe{A}$ \wrt to the gyroscopes frame $\myframe{I}$ by the rotation matrix $\myrotmat{I}{A}$.
The frame $\myframe{G}$ denotes a gravity aligned ($\myvecff{G}{e}{z}=-\mathbf{g}$) inertial frame and is used to express the estimated pose of the sensor system $\mytrafo{G}{I}^k$ and the position of the estimated landmarks $_G{\myvec{l}_m}$.
}
\label{fig:frames_of_ref}
\end{figure}

The following subsections introduce the sensor models for the camera and \ac{IMU}.
An overview of all model parameters is shown in~\reftab{tab:calib_params} and all relevant coordinate frames of the visual and inertial system in~\reffig{fig:frames_of_ref}.

\begin{table}[t]
\centering
\label{tab:calib_params}
\caption{Model parameters of the visual-inertial sensor system.}
\begin{tabular}{@{}llll@{}}
\textbf{Parameter} & \textbf{Symbol} & \textbf{Dim.} &\textbf{Unit}  \\
\toprule
Camera                 &               &     &        \\ \midrule
~~focal length                    & $\calibF$          & $\spaceRealX{2}$    & px  \\
~~principal point                 & $\calibC$          & $\spaceRealX{2}$    & px   \\
~~distortion                      & $\calibW$          & $\spaceReal$     &   -   \\  \addlinespace 

\ac{IMU}                    &  \textbf{~}            &     &  \textbf{~}       \\ \midrule
~~axis misalignment (gyro, accel.) & $\calibMa$, $\calibMg$ & $\spaceRealX{3}$,$\spaceRealX{3}$ & -   \\
~~axis scale (gyro, accel.)        & $\calibSa$, $\calibSg$ & $\spaceRealX{3}$,$\spaceRealX{3}$ & -   \\
~~rotation $\myframe{A}$ \wrt $\myframe{I}$      & $\calibExtAG$      & $\spaceRot$   & -  \\ \addlinespace 

Extrinsics                 &  \textbf{~}            &     &  \textbf{~}    \\ \midrule
~~translation $\myframe{C}$ \wrt $\myframe{I}$       & $\calibExtT$     & $\spaceRealX{3}$    & m   \\
~~rotation $\myframe{C}$ \wrt $\myframe{I}$      & $\calibExtR$       & $\spaceRot$   & -   \\
\bottomrule
\end{tabular}
\end{table}

\subsection{Notation and Definitions}
\label{sec:notation}
A transformation matrix $\mytrafo{A}{B}\in\spaceSE$ takes a vector $\myvecf{p}{B}\in\spaceRealX{3}$ expressed in the frame of reference $\myframe{B}$ into the coordinates of the frame $\myframe{A}$ and can be further partitioned into a rotation matrix $\myrotmat{A}{B}\in\spaceRot$ and a translation vector $\myvecff{A}{p}{AB}\in\spaceRealX{3}$ as follows:
\begin{equation}
\label{eq:trafo}
\begin{bmatrix}
\myvecf{p}{A} \\ 1
\end{bmatrix}
 = \mytrafo{A}{B} \cdot 
\begin{bmatrix}
\myvecf{p}{B} \\ 1
\end{bmatrix}
= 
\begin{bmatrix}
\myrotmat{A}{B} & \myvecff{A}{p}{AB} \\
 \mathbf{0}_{1x3} & 1 
\end{bmatrix}
\cdot 
\begin{bmatrix}
\myvecf{p}{B} \\ 1
\end{bmatrix}
\end{equation}
The unit quaternion $\myquat{A}{B}$ represents the rotation corresponding to $\myrotmat{A}{B}$ as defined in \cite{trawny2005indirect}.
The operator $\mytrafo{A}{B}(\cdot)$ is defined to transform a vector in $\spaceRealX{3}$ from $\myframe{B}$ to the frame of reference $\myframe{A}$ as $\myvecf{p}{A} = \mytrafo{A}{B} \left( \myvecf{p}{B} \right)$ according to \refeq{eq:trafo}.

\subsection{Camera Model}
%
%
A function $f_p(\cdot)$ models the perspective projection and lens distortion effects of the camera.
It maps the $m$-th 3d landmark $\myvecff{C_k}{l}{m}$ onto the image plane of the camera $k$ to yield the 2d image point $\myvec{p}_{k,m}$ as:
\begin{equation}
\begin{aligned}
\myvec{p}_{k,m} = f_p \left( \myvecff{C_k}{l}{m}, \myvec{\theta}_c \right)
\end{aligned}
\end{equation}
where $\myvec{\theta}_c$ denotes the model parameters of the perspective projection function (which we want to calibrate).

In our evaluation setup, we use high-field-of-view cameras as they typically yield more accurate motion estimates \cite{zhang2016benefit}.
As a consequence the camera records a heavily distorted image of the world.
To account for these effects, we augment the pinhole camera model with the field-of-view (FOV) distortion model~\cite{devernay2001straight} to obtain the following perspective projection function:
\begin{equation}
\myvec{p}_{k,m} = f_p(\myvecff{C}{l}{m},\myvec{\theta}_c) = 
\begin{bmatrix}
\beta_r \left( \norm{ \myvec{\overline{p}}_m} \right) \cdot f_x \cdot \overline{p}_x + c_x
\\ 
\beta_r \left( \norm{ \myvec{\overline{p}}_m} \right) \cdot f_y \cdot \overline{p}_y + c_y
\\
\end{bmatrix}
\end{equation}
where $f_{\left(\cdot\right)}$ denotes the focal length, $c_{\left(\cdot\right)}$ the principal point and $\myvec{\overline{p}}$ the 2d projection of a 3d landmark $\myvecff{C}{l}{m}$ in normalized image coordinates as:
\begin{equation}
\myvec{\overline{p}}_{m} = \frac{1}{\myvecff{C}{l}{m}^z} \cdot \begin{bmatrix}
\myvecff{C}{l}{m}^x\\ 
\myvecff{C}{l}{m}^y
\end{bmatrix}
\end{equation}
The function $\beta_r$ models the (symmetric) distortion effects as a function of the radial distance to the optical center as:
\begin{equation}
\beta_r \left( r \right) = \frac{ \arctan{ \left(2 \cdot \tan{\left(\tfrac{w}{2}\right)} \cdot r \right )}}{w \cdot r}
\end{equation}
with $w$ being the single parameter of the FOV distortion model.

%
%
The measurement model for landmark observations expressed in the global frame $\myframe{G}$ (see \reffig{fig:frames_of_ref}) can be written as:
\begin{equation}
\begin{aligned}
\label{eq:landmark_proj}
\myvec{\widetilde{p}}_{k,m} &= f_p\left(\myvecff{C}{l}{m}, \myvec{\theta}_c\right) + \myvec{\eta}_{c} \\
                &= f_p\left(\mytrafo{C}{I}\left(\mytrafo{I}{G}\left( \myvecff{G}{l}{m}\right)\right), \myvec{\theta}_c\right) + \myvec{\eta}_{c}
\end{aligned}
\end{equation}
where $\myvec{\widetilde{p}}_{k,m}$ denotes the projection of the landmark $m$ onto the image plane of the keyframe $k$, $\mytrafo{I}{G}^k$ the pose of the sensor system, $\mytrafo{C}{I}$ the relative pose of the camera \wrt the \ac{IMU} and $\myvec{\eta}_c$ a white Gaussian noise process with zero mean and standard deviation $\sigma_{c}$ as $\myvec{\eta}_c \sim \mathcal{N}(\mathbf 0, \sigma_{c}^2 \cdot \myvec{I}_{2})$.
The full calibration state $\myvec{\theta}_c$ of the camera model can be summarized as:
\begin{equation*}
  \myvec{\theta}_c = \begin{bmatrix}
  {\calibExtR}^T & {\calibExtT}^T & \calibF^T & \calibC^T & \calibW
  \end{bmatrix}^T
\end{equation*}
where the camera-\ac{IMU} relative pose $\mytrafo{C}{I}$ is split into its rotation part $\calibExtR$ and its translation part $\calibExtT$, $\calibF=\begin{bmatrix}f_x & f_y\end{bmatrix}^T$ is the focal length, $\calibC=\begin{bmatrix}c_x & c_y\end{bmatrix}^T$ the principal point and $\calibW$ the distortion parameter of the lens distortion model.

\subsection{Inertial Model}
\label{sec:inertial_model}
The \ac{IMU} considered in this work consists of a (low-cost) \ac{MEMS} 3-axis accelerometer and a 3-axis gyroscope.
As in the work of \cite{li2013high,krebs2012generic,nikolic2016non}, we include the alignment of the non-orthogonal sensing axis and a correction of the measurement scale into our sensor model.
Further, we assume the translation between the accelerometer and gyroscope to be small (single-chip \ac{IMU}) and only model a rotation between the two sensors (as shown in~\reffig{fig:frames_of_ref}).

Considering these effects, we can write the model for the gyroscope measurements $\tilde{\myvec{\omega}}$ as:
\begin{equation}
\tilde{\myvec{\omega}} = \mathbf{T}_g \cdot \myvecff{I}{\omega}{GI} + \myvec{b}_g + \myvec{\eta}_g
\end{equation}
where $\myvecff{I}{\omega}{GI}$ denotes the true angular velocity of the system, $\myvec{T}_a$ a correction matrix accounting for the scale and misalignment of the individual sensor axis (see \refeq{eq:calib_t}), $\myvec{b}_g$ is a random walk process as:
\begin{equation}
\dot{\myvec{b}}_g = \myvec{\eta}_{bg}
\end{equation}
with the zero-mean white noise Gaussian processes being defined as
\begin{align}
\myvec{\eta}_{g} \sim \mathcal{N}(\mathbf 0, \sigma_{g}^2 \cdot \myvec{I}_{3}), \\
\myvec{\eta}_{bg} \sim \mathcal{N}(\mathbf 0, \sigma_{bg}^2 \cdot \myvec{I}_{3}).
\end{align}
%

Similarly, the specific force measurements $\tilde{\myvec{a}}$ of the accelerometer are modeled as:
\begin{equation}
\tilde{\myvec{a}} = \myvec{T}_a \cdot \myrotmat{A}{I} \cdot \myrotmat{I}{G}^k \cdot \left( \myvecff{G}{a}{GI} - _{G}\myvec{g} \right) + \myvec{b}_a + \myvec{\eta}_a
\end{equation}
where $\myvecff{G}{a}{GI}$ is the true acceleration of the sensor system $\myframe{I}$ \wrt to the inertial frame $\myframe{G}$, $\myrotmat{A}{I}$ the relative orientation between the gyroscope and accelerometer frame, $\myrotmat{I}{G}^k$ the orientation of the \ac{IMU} \wrt the inertial frame $\myframe{G}$, $\myvec{T}_a$ is a correction matrix for the scale and misalignment (see \refeq{eq:calib_t}), $_{G}\myvec{g}$ the gravity acceleration expressed in the inertial frame $\myframe{G}$.
The bias process $\myvec{b}_a$ is defined as a random walk process as:
\begin{equation}
\dot{\myvec{b}}_a = \myvec{\eta}_{ba}
\end{equation}
with the zero-mean white noise Gaussian processes being defined as:
\begin{align}
\myvec{\eta}_{a} \sim \mathcal{N}(\mathbf 0, \sigma_{a}^2 \cdot \myvec{I}_{3}), \\
\myvec{\eta}_{ba} \sim \mathcal{N}(\mathbf 0, \sigma_{ba}^2 \cdot \myvec{I}_{3}).
\end{align}

The noise characteristics of the \ac{IMU} $\myvec{\sigma}_{i}=\begin{bmatrix}\sigma_{g} & \sigma_{a} & \sigma_{bg} & \sigma_{ba}\end{bmatrix}^T$ are assumed to have been identified beforehand at nominal operating conditions e.g. using the method described in \cite{woodman2007imu}.
The correction matrix $\myvec{T}_g$ and $\myvec{T}_a$ accounting for the scale and misalignment errors is defined identically for the gyroscope and accelerometer and is partitioned as:
\begin{equation}
  \label{eq:calib_t}
  \mathbf{T}_{\left(\cdot\right)} = \begin{bmatrix}
  s^x_{\left(\cdot\right)} & m^x_{\left(\cdot\right)} & m^y_{\left(\cdot\right)} \\ 
  0 & s^y_{\left(\cdot\right)}  & m^z_{\left(\cdot\right)} \\ 
  0 & 0 & s^z_{\left(\cdot\right)}
  \end{bmatrix}
\end{equation}
where $\myvec{m}_{\left(\cdot\right)}$ denotes the collection of all misalignment and $\myvec{s}_{\left(\cdot\right)}$ all scale factors as:
\begin{equation}
  \myvec{s}_{\left(\cdot\right)} = \left[\begin{smallmatrix}
  s^x_{\left(\cdot\right)} \\ s^y_{\left(\cdot\right)} \\ s^z_{\left(\cdot\right)}
  \end{smallmatrix}\right]
  \quad
  \myvec{m}_{\left(\cdot\right)} = \left[\begin{smallmatrix}
  m^x_{\left(\cdot\right)} \\ m^y_{\left(\cdot\right)} \\ m^z_{\left(\cdot\right)}
  \end{smallmatrix}\right]
\end{equation}

The full calibration state $\myvec{\theta}_i$ of the inertial model can then be summarized as:
\begin{equation}
  \myvec{\theta}_i = \begin{bmatrix}
  \mathbf{s}_g^T & \mathbf{m}_g^T & \mathbf{s}_a^T & \mathbf{m}_a^T & \mathbf{q}_{AI}^T
  \end{bmatrix}^T
\end{equation}
where $\myquat{A}{I}$ describes the rotation of gyroscope frame $\myframe{G}$ \wrt to the accelerometer frame $\myframe{A}$ (with the IMU frame $\myframe{I}$ being defined as the gyroscope frame $\myframe{G}$).

\section{Visual-Inertial Self-Calibration}
\label{sec:vicalib}
In this section, we formulate the self-calibration problem for visual and inertial sensor systems using the sensor models introduced in the previous section.
The derived \ac{ML} estimator makes use of all images and inertial measurements within the dataset to yield a \textit{full-batch} solution.
The motion of the sensor system and the (sparse) scene structure are jointly estimated with the model parameters to achieve self-calibration without the need for a known calibration target (e.g. a chessboard pattern).
The batch estimator will serve as a base to introduce the segment-based calibration which only considers the most informative segments of a trajectory (see \refsec{sec:info_calib}).

\subsection{System State and Measurements}
\label{sec:states}
The self-calibration formulation jointly estimates all keyframe states $\myvec{x}_{k}$, all point landmarks $\myvecff{G}{l}{m}$, the calibration parameters of the camera $\boldsymbol{\theta}_c$ and the \ac{IMU} $\boldsymbol{\theta}_i$ with the keyframe state $\myvec{x}_{k}$ being defined as:
\begin{equation}
\label{eq:keyframe_states}
\myvec{x}_{k}=\begin{bmatrix} {\myquat{G}{I}^{k}}^T & {\myvecff{G}{p}{GI}^{k}}^T & {\myvecff{G}{v}{I}^{k}}^T & {\myvec{b}_{a}^{k}}^T & {\myvec{b}_{g}^{k}}^T \end{bmatrix}^T
\end{equation}
where ${\myquat{G}{I}}^{k}$ and $\myvecff{G}{p}{GI}^{k}$ define the pose of the sensor system at timestep $k$, $\myvecff{G}{v}{I}^{k}$ the velocity of the system and $\myvec{b}_{\left(\cdot\right)}^{k}$ the bias of the gyroscope and accelerometer.

To simplify further notations, we collect all states of the problem in the following vectors:
\begin{equation}
\myvec{\hat{x}}_{0..K} = \left[\begin{smallmatrix} \myvec{\hat x}_0 \\ \vdots \\ \myvec{\hat x}_K \end{smallmatrix}\right]
\quad
\myvecf{\hat{l}}{G}_{0..M} = \left[\begin{smallmatrix} \myvecff{G}{\hat l}{0} \\ \vdots \\ \myvecff{G}{\hat l}{M} \\ \end{smallmatrix} \right]
\quad
\myvec{\hat{\theta}} = \left[\begin{smallmatrix} {\myvec{\hat{\theta}}_c} \\ {\myvec{\hat{\theta}}_i} \end{smallmatrix}\right]
\end{equation}
where $K$ is the total number of keyframes and $M$ the number of landmarks.
Additionally, the vector $\boldsymbol{\hat{\pi}}_{K,M}$ stacks all estimated states as:
\begin{equation}
\begin{aligned}
\boldsymbol{\hat{\pi}}_{K,M} &= \begin{bmatrix} {\myvec{\hat{x}}_{0..K}}^T & {\myvecf{\hat{l}}{G}_{0..M}}^T & {\myvec{\hat{\theta}}}^T \end{bmatrix}^T
\end{aligned}
\end{equation}

Further, we define the collection $U$ to contain all \ac{IMU} measurements and $Z$ all 2d landmark observations of the camera as:
\begin{equation}
\begin{aligned}
\mathbf{U}&=\{\myvec{u}_k | k \in [0,K-1]\} \\
\mathbf{Z}&=\{\myvec{p}_{k,m} | k \in [0, K], m \in [0, M(k)]\} 
\end{aligned}
\end{equation}
where $\myvec{u}_k$ is the set of all accelerometer and gyroscope measurements between the keyframes $k$ and $k+1$ and $\myvec{p}_{k,m}$ the 2d measurement of the $m$-th landmark seen from the $k$-th keyframe and $K$ and $M$ denote the number of keyframes and landmarks respectively.

\subsection{State Initialization using VIO}
\label{sec:state_init_vio}
A vision front-end tracks sparse point features between consecutive images and rejects potential outliers based on geometrical consistency using a perspective-n-point algorithm in a RANSAC scheme.
The resulting feature tracks and the \ac{IMU} measurements are processed by an \ac{EKF} which is loosely based on the formulation of \cite{mourikis2007multi,li2014high} but with various extension to increase robustness and accuracy.
The filter recursively estimates all keyframe states $\myvec{x}_{0..K}$ and landmark positions $\myvecf{l}{G}_{0..M}$.
The calibration states are not estimated by this filter except for the camera-to-\ac{IMU} relative pose (camera extrinsics).
However, for the initialization of the calibration problem, we only use the keyframe states (pose, velocity, biases) and the most recent estimate of the camera-to-\ac{IMU} extrinsics.
The landmark states are initialized by triangulation using the poses estimated by the \ac{EKF} filter.

It is important to note that the filter needs sufficiently good calibration parameters in order to run properly and provide accurate initial estimates.
In our experience, it is sufficient for most single-chip \ac{IMU}s to initialize their intrinsic calibration to a nominal value (unit scale, no misalignment).
However, a complete self-calibration may be difficult if no priors are available for the camera intrinsics.
In this case, a specialized calibration method should be used beforehand e.g. \cite{rehder2016extending,zhang2000flexible}.

\subsection{ML-based Self-Calibration Problem}
\label{sec:batch_calib}
We use the framework of \ac{ML} estimation to jointly infer the state of all keyframes $\myvec{\hat{x}}_{0..K}$, landmarks $\myvecf{\hat{l}}{G}_{0..M}$ and calibration parameters $\myvec{\hat{\theta}}$ using all available measurements $U$ of the \ac{IMU} and the 2d measurements $Z$ of the point landmarks extracted from the camera images.
\begin{figure}[t]
  \centering
  \includegraphics[width=7.0cm]{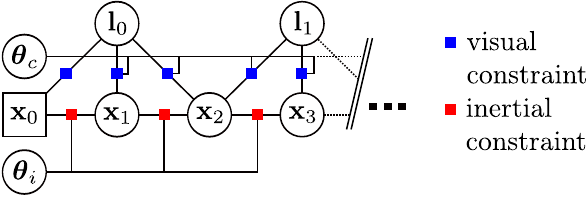}
  \caption{%
  Batch calibration problem shown in factor-graph representation:
  the problem contains keyframe states $\myvec{x}_k$ (pose, velocity, gyroscope and accelerometer biases), the calibration states for the \ac{IMU} $\boldsymbol{\theta}_i$ and the camera $\boldsymbol{\theta}_c$ and the landmarks $\myvecff{}{l}{m}$. 
  Two types of factor are used: (red) inertial constraints $g_k^{imu}\left(\myvec{x}_k,\myvec{x}_{k+1},\boldsymbol{\theta}_i,\myvec{u}_k\right)$ based on the integrated \ac{IMU} measurements; (blue) landmark reprojection factors $g_{k,m}^{cam}\left(\myvec{x}_k,\myvecff{}{l}{m},\myvec{p}_{k,m}\right)$ modeling the feature observations (measurements of a landmark projection) observed by the camera.
  Additionally, the unconstrained directions of the first keyframe state, namely the global position $\myvecff{G}{p}{GI}^0$ and the rotation around the gravity vector $\myquat{G}{I}^0$ (z-axis of frame $\myframe{G}$) are fixed to zero (denoted by the square).  
  }
\label{fig:factor_graph_batch}
\end{figure}
A factor graph representation of the visual-inertial self-calibration formulation is shown in \reffig{fig:factor_graph_batch}.
The problem contains two types of factor: the visual factor $g^{cam}_{k,m}$ models the projection of the landmark $m$ onto the image plane of the keyframe $k$ and the inertial factor $g^{imu}_{k}$ forms a differential constraint between two consecutive keyframe states $x_k$ and $x_{k+1}$ (pose, velocity, bias).
The \ac{ML} estimate $\boldsymbol{\hat\pi}_{ML}$ is obtained by a maximization of the corresponding likelihood function $p(\boldsymbol{\pi} | \mathbf{Z},\mathbf{U})$.
When assuming Gaussian noise for all sensor models (see \refsec{sec:vi_system}), the \ac{ML} solution can be approximated by solving the (non-linear) least-squares problem with the following objective function $S(\boldsymbol{\pi})$:
\begin{equation}
\begin{aligned}
S(\boldsymbol{\pi}) = &\sum_{k=0}^{K} \sum_{m}^{M(k)} {\myvec{e}_{k,m}^{cam}}^T \myvec{W}_{k,m}^{cam} \myvec{e}_{k,m}^{cam} \\
+ &\sum_{k=0}^{K-1} {\myvec{e}_k^{imu}}^T \myvec{W}_k^{imu} \myvec{e}_k^{imu}
\end{aligned}
\end{equation}
where $K$ denotes the number of keyframes, $M(k)$ the set of landmarks off from keyframe $k$, $\myvec{e}_{k,m}^{cam}$ the reprojection error of the $m$-th point landmark of observed from the $k$-th keyframe and $\myvec{e}_k^{imu}$ denotes the inertial constraint error between two consecutive keyframe states $k$ and $k+1$ as a function of integrated \ac{IMU} measurements.
The terms $\myvec{W}_{k,m}^{cam}$ and $\myvec{W}_k^{imu}$ denote the inverse of the error covariance matrices: keypoint measurement and the integrated \ac{IMU} measurement covariance respectively.
The reprojection error $\myvec{e}_{k,m}^{cam}$ is defined as:
\begin{equation}
    \myvec{e}_{k,m}^{cam} = \myvec{p}_{k,m} - \myvec{\widetilde{p}}_{k,m} \left(\mytrafo{C}{I},\mytrafo{I}{G}^k,\myvecff{G}{l}{m},\myvec{\theta}_c\right)
\end{equation}
where $\myvec{p}_{k,m}$ is the 2d measurement of the projection of the landmark $m$ into camera $k$ and $\myvec{\widetilde{p}}_{k,m}$ its prediction as defined in \refeq{eq:landmark_proj}.
The inertial error $\myvec{e}_k^{imu}$ is obtained by integrating the continuous equations of motion using the sensor models described in \refsec{sec:inertial_model} and is based on the method described in \cite{li2014high}.
The non-linear objective function $S(\boldsymbol{\pi})$ is minimized using numerical optimization methods.
In our implementation, we use the Levenberg-Marquardt implementation of the Ceres framework \cite{ceres_solver}.

\section{Self-Calibration using Informative Motion Segments}
\label{sec:info_calib}
%
%
In this section, we propose a method to identify informative segments in a calibration dataset and a modified formulation for estimating calibration parameters based on a set of segments.
First, the method can be used to sparsify a dataset and consequently reduce the complexity of the optimization problem.
And second, a complete calibration dataset can be built over time by accumulating informative segments from multiple sessions,
thus enabling the calibration of even weakly observable parameters by collecting exciting motion that occurs eventually.
It is important to note that the proposed method is presented on the use-case of visual-inertial calibration but it can be applied to arbitrary calibration problems.

\begin{figure}[t]
\centering
\includegraphics[width=8cm]{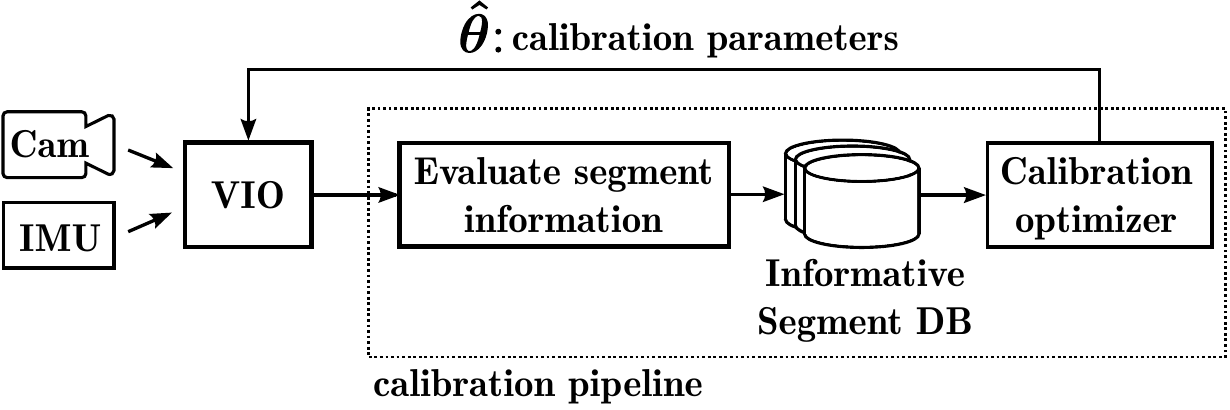}
  \caption{High-level overview of the modules and data flows of the proposed method: (1) motion estimates from \ac{VIO} are used to identify informative motion segments, (2) the most informative segments are maintained in a database for later calibration and (3) an \ac{ML}-based calibration is triggered once enough data has been collected to update the sensor calibration.}
\label{fig:arch}
\end{figure}

\subsection{Architecture}
%
%
A high-level overview of the modules and data-flows is shown in \reffig{fig:arch}.
The proposed method is intended to be run in parallel to an existing visual-inertial motion estimation system.
The \ac{VIO} implementation used in this work is described in \refsec{sec:state_init_vio} but it is important to note that the method is not tied to a particular motion estimation framework.
The keyframe and landmarks states estimated by the \ac{VIO} module are partitioned into segments.
In a next step, the information content of each segment \wrt the calibration parameters is evaluated using an efficient information theoretic metric.
A database maintains the most informative segments of the trajectory and a calibration is triggered once enough data has been collected.
This algorithm is summarized in~\refalg{alg:calib_algo} and explained in more details in the following sections.

\begin{algorithm}[!ht]
 \definecolor{commentcolor}{RGB}{150,0,0}
 \footnotesize
 \KwIn{Initial calibration: $\boldsymbol{\hat\theta}_{init}$}
 \KwOut{Updated calibration: $\boldsymbol{\hat\theta}$}
 \SetKwFunction{waitsensor}{WaitForNewSensorData}
 \SetKwFunction{runvio}{RunVIO}
 \SetKwFunction{collect}{CollectNKeyframes}
 \SetKwFunction{evalinfo}{EvaluateSegmentInformation}
 \SetKwFunction{adddb}{UpdateDatabase}
 \SetKwFunction{dbfull}{EnoughSegmentsInDatabase}
 \SetKwFunction{runcalib}{RunOptimization}
 \SetKwFunction{getdb}{GetAllSegmentsFromDatabase}
 \SetKwFor{Loop}{Loop}{}{EndLoop}

 \BlankLine
 \Loop{} { 
  \textcolor{commentcolor}{// Initialize motion segments of size N from \ac{VIO} output.}\\
  $\mathcal{S}_{i} \leftarrow \{\}$  \\
  \Repeat{$dim(\mathcal{S}_{i}) == N$}{
    $data$ = \waitsensor{} \\
    $\mathbf{\hat x}_j$, $\mathbf{\hat l}_j \leftarrow$ \runvio{$data$, $\hat{\theta}_{init}$} \textcolor{commentcolor}{// \refsec{sec:state_init_vio}}\\ 
    $\mathcal{S}_{i} \leftarrow \mathcal{S}_{i} \cup (\mathbf{\hat x}_j$, $\mathbf{\hat l}_j)$
  }
  \BlankLine
  $H\left(\boldsymbol{\theta}\right) \leftarrow$ \evalinfo{$\mathcal{S}_i$} \textcolor{commentcolor}{// \refsec{sec:eval_segment_info}}\\ 
  \adddb{$\mathcal{S}_{i}$, $H\left(\boldsymbol{\theta}\right)$} \textcolor{commentcolor}{// \refsec{sec:info_db}}\\    
  \If{\dbfull{}} {
    $\mathcal{\mathbf{S}}_{info} \leftarrow$ \getdb{} \\
    $\boldsymbol{\hat \theta} \leftarrow$ \runcalib{$\mathcal{\mathbf{S}}_{info}$}   \textcolor{commentcolor}{// \refsec{sec:segment_problem}} \\
	\KwRet $\boldsymbol{\hat \theta}$
  }
  $i \leftarrow i + 1$
 }
 
 \caption{Self-calibration on informative motion segments.}
 \label{alg:calib_algo}
\end{algorithm}

\subsection{Evaluating Information Content of Segments}
\label{sec:eval_segment_info}
%
%
The continuous stream of keyframe $\myvec{\hat{x}}_k$ (pose, velocity, bias) and landmark states $\myvecff{G}{\hat{l}}{m}$, estimated by the \ac{VIO}, is partitioned into motion segments.
The $i$-th segment $\mathcal{S}_i$ is made up by the $N$ consecutive keyframes $\mathcal{\boldsymbol{\hat{X}}}_i=\myvec{\hat{x}}_{\left(i \cdot N\right)..\left(\left(i+1\right) \cdot N - 1\right)}$  and the set of landmarks $\mathcal{\boldsymbol{\hat{L}}}_i$ observed from this segment.

%
%
We propose to use information metrics that only consider the constraints within each segment to evaluate the information content \wrt the calibration parameters $\boldsymbol{\theta}$.
Using such an information metric which is independent of all other segments makes its evaluation very efficient at the cost of neglecting cross-terms coming from other segments such as loop-closure constraints.
However, the neglected constraints can be re-introduced and considered during the calibration.
Thus, this assumption only affects the selection of informative segments and potentially leads to a conservative estimate of the actual information but should not bias the calibration results.

%
%
To quantify the information content of the $i$-th segment ${\mathcal{S}}_i$, we recover the marginal covariance $\boldsymbol{\Sigma}_{\theta}^{\mathcal{S}_i} = \operatorname{Cov} \left [p(\boldsymbol{\theta} | \myvec{U}_i,\myvec{Z}_i) \right]$ of the calibration parameters $\boldsymbol{\theta}$ given all the constraints within the segment.
For this, we first approximate the covariance $\boldsymbol{\Sigma}_{\mathcal{X}\mathcal{L}\theta}^{\mathcal{S}_i}$ over all segment states using the \textit{Fisher Information Matrix} as:
\begin{equation}
\label{eq:segment_cov_fim}
\boldsymbol{\Sigma}_{\mathcal{X}\mathcal{L}\theta}^{\mathcal{S}_i} = \operatorname{Cov}\left[ p(\mathcal{\boldsymbol{X}}_i, \mathcal{\boldsymbol{L}}_i, \boldsymbol{\theta} | \myvec{U}_i,\myvec{Z}_i) \right] = (\myvec{J}_i^T \myvec{G}_i^{-1} \myvec{J}_i)^{-1} 
\end{equation}
The matrix $\myvec{J}_i$ represents the stacked Jacobians of all error terms $\myvec{e}_k$ and $\myvec{G}_i$ the stacked error covariances $\mathbf{W}_k$ corresponding to the errors terms as:
\begin{equation}
  \myvec{J}_i = \left[\begin{smallmatrix}
   {\frac{\partial \myvec{e}_0}{\partial \myvec{\Pi}_i}}    \\     \vdots       \\     \frac{\partial \myvec{e}_K^T}{\partial \myvec{\Pi}_i}
  \end{smallmatrix}\right],
\quad
    \myvec{G}_i :=  diag \{\myvec{W}_0 \hdots , \myvec{W}_K\}
\end{equation}
where $\myvec{\Pi}_i = \left[ \mathcal{\boldsymbol{X}}_i, \mathcal{\boldsymbol{L}}_i, \boldsymbol{\theta} \right]$ denotes the collection of all states within the segment $i$ and $K$ the number of errors terms within the segment $i$.
Further, the state ordering is chosen such that the rightmost columns of $\boldsymbol{\Sigma}_{\mathcal{X}\mathcal{L}\theta}^{\mathcal{S}_i}$ correspond to the states of the calibration parameters $\boldsymbol{\theta}$.

A rank-revealing QR decomposition is used to obtain $\mathbf{Q_i}\mathbf{R_i}=\mathbf{L}_i\mathbf{J}_i$ with $\mathbf{G}_i^{-1}=\mathbf{L}_i^T\mathbf{L}_i$ being the Cholesky decomposition of the error covariance matrix.
The \refeq{eq:segment_cov_fim} can then be rewritten as
\begin{equation}
\label{eq:cholesky}
\boldsymbol{\Sigma}_{\mathcal{X}\mathcal{L}\theta}^{\mathcal{S}_i} = 
(\myvec{R}_i^T\myvec{R}_i)^{-1} = 
\begin{bmatrix}
 \boldsymbol{\Sigma}_{\mathcal{X}\mathcal{L}}^{\mathcal{S}_i} &
 \boldsymbol{\Sigma}_{\mathcal{X}\mathcal{L},\theta}^{\mathcal{S}_i} \\ 
{\boldsymbol{\Sigma}_{\mathcal{X}\mathcal{L},\theta}^{\mathcal{S}_i}}^T & 
\boldsymbol{\Sigma}_{\theta}^{\mathcal{S}_i} \\ 
\end{bmatrix}
\end{equation}
As $\myvec{R}_i$ is an upper-triangular matrix, we can obtain the marginal covariance $\boldsymbol{\Sigma}_{\theta}^{\mathcal{S}_i}$ efficiently by back-substitution.

%
%
In a next step, we normalize the marginal covariance $\boldsymbol{\Sigma}_{\theta}^{\mathcal{S}_i}$ to account for different scales of the calibration parameters with:
\begin{equation}
\label{eq:marginal_normalization}
\overline{\boldsymbol{\Sigma}}_{\theta}^{\mathcal{S}_i} = \operatorname{diag}(\boldsymbol{\sigma}_{ref})^{-1} \cdot \boldsymbol{\Sigma}_{\theta}^{\mathcal{S}_i} \cdot \operatorname{diag}(\boldsymbol{\sigma}_{ref})^{-1}
\end{equation}
where $\boldsymbol{\sigma}_{ref}$ is the expected standard deviation that has been obtained empirically from a set of segments from various datasets.
It is important to note, that $\boldsymbol{\sigma}_{ref}$ depends on the sensor setup (e.g. focal length, dimensions, etc.) and should either be re-evaluated for each setup or a normalization based on nominal calibration parameters should be performed.

%
%
We can now define different information metrics based on the normalized marginal covariance $\overline{\boldsymbol{\Sigma}}_{\theta}^{\mathcal{S}_i}$.
These metrics will be used to compare segments based on their information content \wrt the calibration parameters $\boldsymbol{\theta}$.
They are defined such that a lower value corresponds to more information.
In this work, we will investigate the three most common information-theoretic metrics from optimal design theory:
\subsubsection{A-Optimality}
This criterion seeks to minimize the trace of the covariance matrix which results in a minimization of the mean variance of the calibration parameters.
The corresponding information metric is defined as:
\begin{equation}
H_{Aopt}^i = \operatorname{trace}\left( \overline{\boldsymbol{\Sigma}}_{\theta}^{\mathcal{S}_i} \right)
\end{equation}

\subsubsection{D-Optimality}
Minimizes the determinant of the covariance matrix which results in a maximization of the differential Shannon information of the calibration parameters.
\begin{equation}
H_{Dopt}^i = \operatorname{det}\left( \overline{\boldsymbol{\Sigma}}_{\theta}^{\mathcal{S}_i} \right)
\end{equation}
It is interesting to note that this criterion is equivalent to the minimization of the differential entropy $H_{e}^i(\boldsymbol\theta)$ which for Gaussian distributions is defined as:
\begin{equation}
\begin{aligned}
H_{e}^i\left(\boldsymbol{\theta}\right) &= -\int _{-\infty }^{\infty } \cdots \int _{-\infty }^{\infty } \bar p_{\boldsymbol{\theta}}(\boldsymbol{\theta})\ln \bar p_{\boldsymbol{\theta}}(\boldsymbol{\theta})\,d\boldsymbol{\theta} \\
&={\frac {1}{2}}\ln \left((2\pi e)^{k}\cdot \operatorname{det}\left(\overline{\boldsymbol{\Sigma}}_{\theta}^{\mathcal{S}_i}\right)\right) 
\end{aligned}
\end{equation}
where $\bar p_{\boldsymbol{\theta}}(\boldsymbol{\theta})=\bar p(\boldsymbol{\theta} | \mathbf{U}_i,\mathbf{Z}_i)$ is the normalized normal distribution of $\boldsymbol{\theta}$ and $k$ the dimension of this distribution.

\subsubsection{E-Optimality}
This design seeks to minimize the maximal eigenvalue of the covariance matrix with the metric being defined as:
\begin{equation}
H_{Eopt}^i = \operatorname{max}\left(\operatorname{eig}\left( \overline{\boldsymbol{\Sigma}}_{\theta}^{\mathcal{S}_i} \right)\right)
\end{equation}

\begin{figure*}[t]
\centering
\includegraphics[width=17cm]{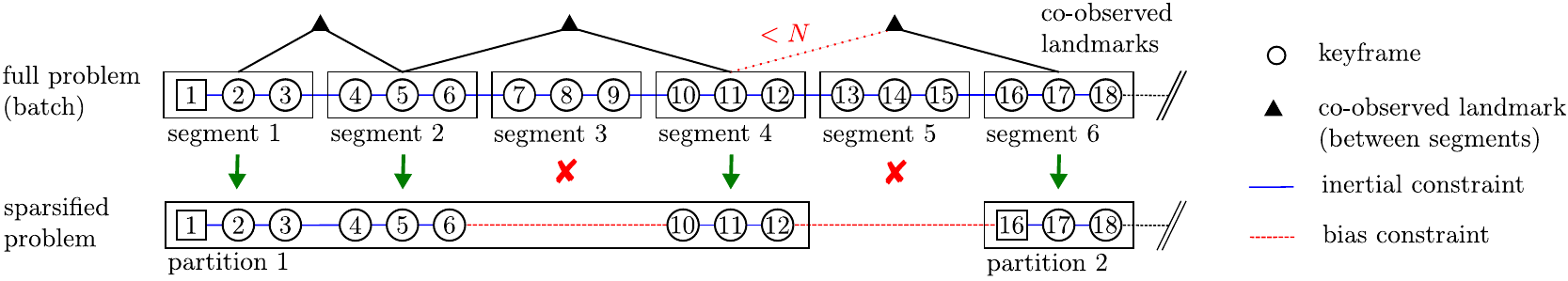}
  \caption{%
  The segment calibration problem only includes the most informative segments of the motion trajectory (keyframes) estimated by the \ac{VIO} (upper graph).
  A constraint on the bias evolution is introduced where non-informative segments have been removed (red cross) while the pose and velocity remain unconstrained.
  Additionally, the segments are partitioned such that each partition co-observes less than $N$ landmarks of other partitions.
  Consequently, the unobservable modes of each partition, namely the rotation around gravity and the global position, are held constant during the optimization for exactly one keyframe of the partition (marked with a square).
  }
\label{fig:segment_partitioning}
\end{figure*}

\subsection{Collection of Informative Segments}
\label{sec:info_db}
We want to maximize the information contained within a fixed-sized budget of segments.
For this reason, we maintain a database with a maximum capacity of $N$ segments retaining only the most informative segments of the trajectory.
The information metric will be used to decide which segments are retained and which are rejected such that the sum over the information metric of all segments in the database is minimized.
Such a decision scheme will ensure that the accumulated information on the calibration parameter $\boldsymbol{\theta}$ is increasing over time while the number of segments remains constant.
Therefore, an upper bound on the calibration problem complexity can be guaranteed.
However, it is important to note that the sum of information metrics is only a conservative approximation of the total information content for two reasons:
First, the information metric is only a scalar and therefore no directional information is available.
Second, the information metrics neglect any cross-terms to other segments and thus underestimates the true information.

\subsection{Segment Calibration Problem}
\label{sec:segment_problem}
%
%
The segment-based calibration differs from the batch estimator introduced in \refsec{sec:vicalib} in that it only contains the most informative segments of a (multi-session) dataset. 
The removal of trajectory segments from the original problem leads to two main challenges.

%
%
First, the time difference between two (temporally neighboring) keyframes could become arbitrarily large when non-informative keyframes have been removed in-between.
An illustration of such a dataset with a temporal gap due to the keyframe removal is shown in \reffig{fig:segment_partitioning} (between keyframe 6/10 and 12/16).
In this case, we only constrain the bias evolution between the two neighboring keyframes using a random walk model described in \refsec{sec:inertial_model} and no constraints are introduced for the remaining keyframe states (pose, velocity).

%
%
Second, the removal of non-informative trajectory segments often creates partitions of keyframes that are neither constrained to other partitions through (sufficient) shared landmark observations nor through inertial constraints.
Each of these partitions can be seen as a (nearly) independent calibration problem that only shares the calibration states with other partitions.
Assuming non-degenerate motion and sufficient visual constraints, each of these partitions contains the 2 structurally unobservable modes of the visual-inertial optimization problem namely the rotation around the gravity vector (yaw in global frame) and the global position.
These modes are eliminated from the optimization by keeping them constant for exactly one keyframe in each of the partitions to achieve efficient convergence of the iterative solvers.

%
%
We identify the partitions based on the co-visibility of landmarks and the connectivity through inertial constraints.
An overview of the algorithm is shown in~\refalg{alg:union_find}.
In a first step, all segments that are direct temporal neighbors, and thus connected through inertial constraints, are joined into larger segments (e.g. segment 1 and 2).
In a next step, we use a union-find data structure to iteratively partition the joined segments into disjoint sets (partitions) such that the number of co-observed landmarks between the partitions lies below a certain threshold.
At this point, all keyframes within a partition are either constrained through inertial measurements or through sufficient landmark co-observations \wrt each other.
It is important to note that degenerate landmark configurations are still possible using such a heuristic metric.
However, an error will only influence the convergence rate of the incremental optimization but should not bias the calibration results.

\begin{algorithm}[t]
 \footnotesize
 \KwIn{Set of motion segments $\mathcal{\mathbf{S}} = \{\mathcal{S}_0, ...,  \mathcal{S}_K\}$}
 \KwIn{Max. co-observed landmarks between partitions $N$}
 \KwResult{Set of motion segment partitions $\mathcal{P}$ }
 \SetKwFunction{Union}{MergePartitions}
 \SetKwFunction{Count}{CountSharedLandmarks}
 \BlankLine
 $\mathcal{\mathbf{P}} \leftarrow \{\}$\; \\
 \ForEach{$\mathcal{S}_k \in \mathcal{\mathbf{S}}$}  {
    $\mathcal{\mathbf{C}} \leftarrow \{\{\mathcal{S}_k\}\}$\; \\
    \ForEach{$p \in \mathcal{\mathbf{P}}$} {
        \If{\Count($p$, $\mathcal{S}_k$) $> N$} {
          $\mathcal{\mathbf{C}} \leftarrow \mathcal{\mathbf{C}} \cup \{p\}$ \;
		}
    }
   $p_{\mathcal{C}}$ $\leftarrow$ \Union{$\mathcal{\mathbf{C}}$}\; \\ 
   $\mathcal{\mathbf{P}} \leftarrow \left( \mathcal{\mathbf{P}} \setminus \mathcal{\mathbf{C}} \right) \cup \{p_{\mathcal{C}}\}$ \;
 }
 \caption{Partitioning segments on landmark co-visibility}
 \label{alg:union_find}
\end{algorithm}

\section{Experimental Setup}
\label{sec:experiments}
This section introduces the experiments, datasets, and hardware used to evaluate the proposed method.
The results are discussed in the next section.

\subsection{Single-/Multi Session Database}
\label{sec:single_multi}
We evaluate the proposed method using two different strategies to maintain informative segments in the database.
Each strategy is investigated using a set of multi-session datasets and discussed along  a suitable use-case:

\subsubsection{Single-session Database: Observability-aware Sparsification of Calibration Datasets} 
\label{sec:llcal_mode_single_session}
Each session starts with an empty segment database and the $N$ most informative segments from this single session are kept.
After each session, a segment-based calibration is performed using all the segments in the database and the calibration parameters are updated for use in the next session.
This strategy can be seen as an observability-aware sparsification method for calibration datasets.
It is well suited for infrequent and long sessions (e.g. \textit{navigation use-case} with lots of still phases) where batch calibration over the entire dataset would be too expensive and data selection is necessary.

\subsubsection{Multi-session Database: Accumulation of Information over Time}
\label{sec:llcal_mode_multi_session}
The multi-session strategy does not reset the database between sessions and the most informative segments are collected from multiple consecutive sessions.
In contrast to the single-session strategy, it is particularly suited for frequent and short sessions; for example in an \textit{AR/VR use-case} where a user performs many short session over a short period of time.
It accumulates information from multiple sessions and thus enables the calibration of weakly observable modes which might not be sufficiently excited in a single session.

\begin{figure}[t]
\centering
\includegraphics[width=8cm]{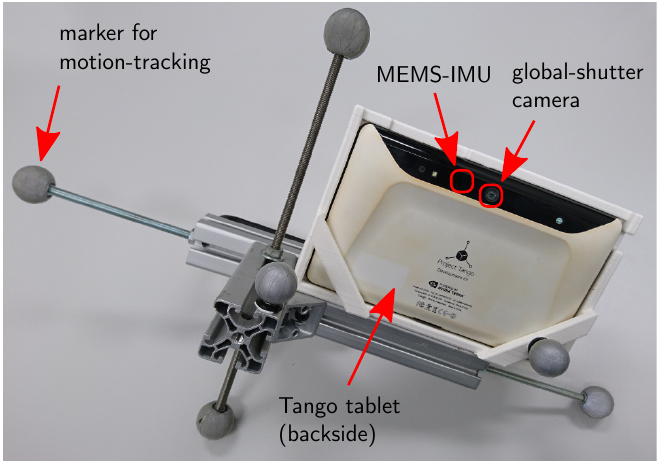}
  \caption{
  The Google Tango tablet used for the dataset collection is equipped with markers for external pose tracking by a Vicon motion-capture system.
  The tablet contains a sensor suite specifically designed for motion tracking including a high field-of-view camera and a single-chip \ac{MEMS} \ac{IMU}.
  }
\label{fig:tango_tablet}
\end{figure}

\begin{table}[t]
\caption{%
Datasets used for the evaluation.
All datasets have been recorded using a Google Tango tablet as shown in \reffig{fig:tango_tablet}.
}
\label{tab:datasets}
\resizebox{\columnwidth}{!}{%
\begin{tabular}{@{}llllll@{}}
\toprule
         &   \textbf{avg. length}   & \textbf{avg. linear} /   &  \\
\textbf{dataset}  &   \textbf{duration}      & \textbf{angular vel.}    & \textbf{description} \\ \midrule
\textbf{AR/VR use-case:} & & & \\
~~office room      &   23.8 m       &   0.20 m/s      & well-lit, good           \\
~~~~(5 sessions)   &   117.3 s      &   20.3 deg/s    & texture                  \\ 
~~class room       &   37.4 m       &   0.29 m/s      & well-lit, open space,    \\
~~~~(5 sessions)   &   122.9 s      &   29.62 deg/s   & good texture             \\ \addlinespace
\textbf{Navigation use-case:} & & & \\
~~parking garage   &   168.4 m      &   0.57 m/s      & dark, low-texture        \\
~~~~(3 sessions)   &   305.0 s      &   20.51 deg/s   & walls, open space        \\ 
~~office building  &   164.8 m      &   0.55 m/s      & well-lit, good           \\   
~~~~(3 sessions)   &   295.6 s      &   23.12 deg/s   & texture, corridors       \\ \addlinespace
\textbf{Evaluation datasets:} & & & \\
~~Vicon room      &   59.7 m        &   0.49 m/s      & motion-capture data,     \\
~~~~(15 sessions) &   114.1 s       &   42.95 deg/s   & well-lit                 \\ \bottomrule
\end{tabular}
}%
\end{table}

\begin{figure}[t]
\centering
\subfigure[AR/VR: office room]{\includegraphics[width=0.22\textwidth]{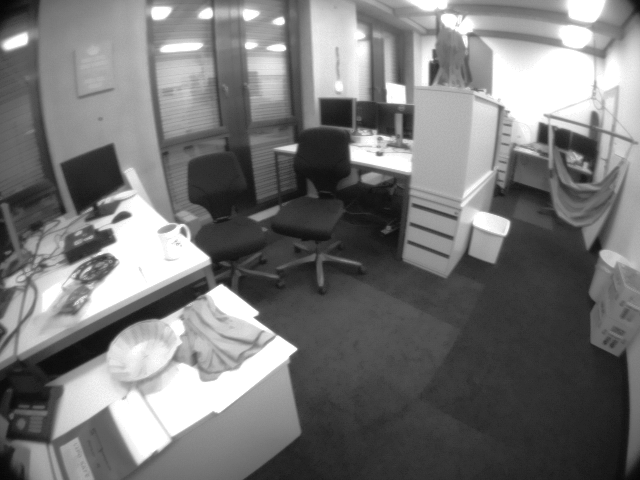}}
\subfigure[AR/VR: class room]{\includegraphics[width=0.22\textwidth]{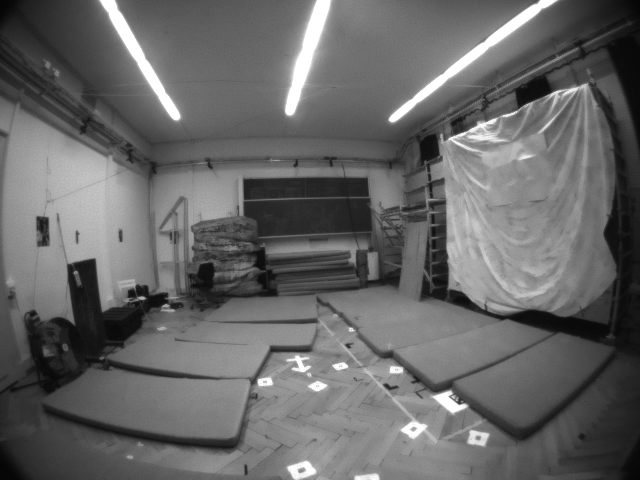}}
\subfigure[NAV: parking garage]{\includegraphics[width=0.22\textwidth]{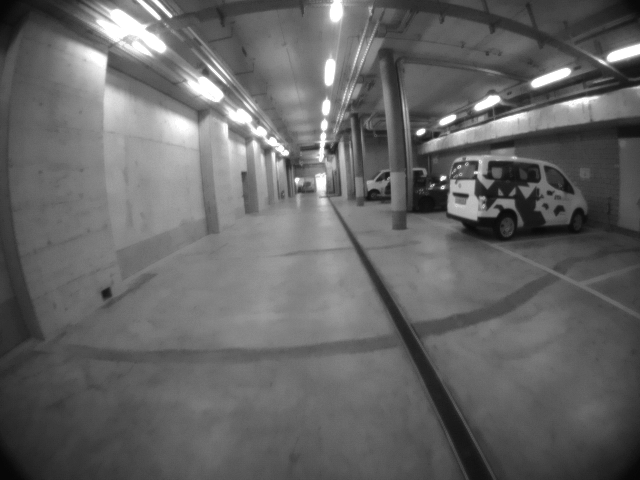}}
\subfigure[NAV: office building]{\includegraphics[width=0.22\textwidth]{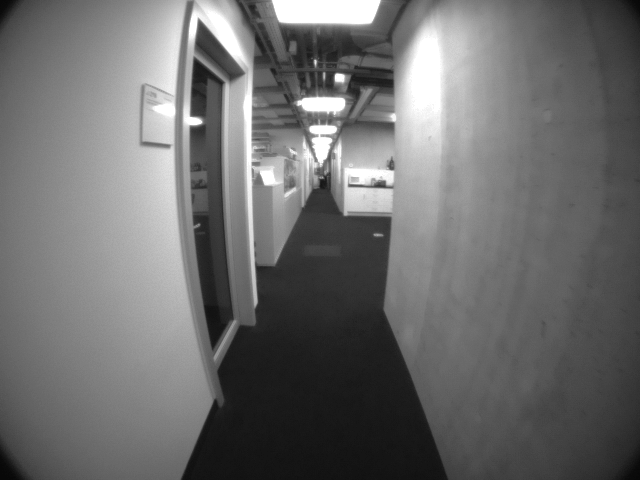}}
\caption{The four different environments in which the calibration datasets have been recorded. The images were taken by the motion tracking camera of the Tango tablet.} \label{fig:dataset_images}
\end{figure}

\subsection{Datasets and Hardware}
\label{sec:datasets}
All datasets were recorded using a Google Tango tablet as shown in \reffig{fig:tango_tablet}.
This device uses a high-field-of-view global shutter camera (10~Hz) and a single-chip \ac{MEMS} \ac{IMU} (100~Hz).
The measurements of both sensors are time-stamped in hardware on a single clock for an accurate synchronization.
Additionally, the sensor rig is equipped with markers for external tracking by a Vicon motion capture system.
All datasets were recorded on the same device, in a short period of time and while trying to keep the environmental factors constant (e.g. temperature) to minimize potential variations of the calibration parameters across the datasets and sessions.

We have collected datasets representative for each of the two use-cases introduced in the previous section in different environments (office, class room, and garage).
These datasets consist of multiple sessions that will be used to obtain a calibration using the proposed method.
Right after recording the calibration datasets, we have collected a batch of $15$ evaluation datasets with motion capture ground-truth.
These datasets are used to evaluate the motion estimation accuracy that can be achieved using the obtained calibration parameters.
An overview of all datasets and their characteristics is shown in \reftab{tab:datasets} and \reffig{fig:dataset_images}.

While recording the calibration datasets, we tried to achieve the following characteristics representative for the two use-cases:
\subsubsection{AR/VR use-case}
We collected datasets that mimic an AR/VR use-case to evaluate whether we can accumulate information from multiple-sessions (multi-session database strategy).
Characteristic of this use-case, the datasets consists of multiple short sessions restricted to a small indoor space (single room), containing mostly fast rotations, only slow and minor translation and stationary phases. 
Two datasets have been recorded in a \textit{class} and \textit{office} room each containing $5$ sessions that are $2$ min long.

\subsubsection{Navigation use-case}
In contrast to the AR/VR use-case, the navigation sessions contain mostly translation over an area of multiple rooms and only slow rotations but also contain stationary and rotation-only phases.
Datasets have been recorded in two locations: \textit{garage} and \textit{office} - each contains $3$ sessions with a duration of $5$ min.
These datasets will be used to evaluate the observability-aware sparsification (single-session database strategy).

\subsection{Evaluation Method}
For performance evaluation, we calibrate the sensor models on each session of the dataset in temporal order where we use the calibration parameters obtained from the previous session as initial values.
The first session uses a nominal calibration consisting of a relative pose between camera and \ac{IMU} from CAD values, nominal values for the \ac{IMU} intrinsics (unit scale factors, no axis misalignment) and camera intrinsics.

This calibration scheme is performed for all datasets and for both of the database strategies to obtain a set of calibration parameters for each session.
The quality of the obtained calibration parameters is then evaluated using the following methods:

\subsubsection{Motion estimation performance}
\label{sec:exp_vicon}
As the main objective of our work is to calibrate the sensor system for ego-motion estimation, we use the accuracy of the motion estimation (based on our calibrations) as the main evaluation metric.
We run all $15$ evaluation datasets for each set of calibration parameters and evaluate the accuracy of the estimated trajectory against the ground-truth from the motion-capture system. 

The motion estimation error is obtained by first performing a spatio-temporal alignment of the estimated and the ground-truth trajectory.
Second, a relative pose error is computed at each time-step between the two trajectories.
To compare different runs, we use the \ac{RMSE} calculated over all the relative pose errors.

\subsubsection{Parameter repeatability}
We only evaluate the parameter repeatability over different calibrations of the same device as no ground-truth for the calibration parameters is available.
We have recorded all dataset close in time while keeping the environmental conditions (e.g. temperature) similar and avoiding any shocks to minimize potential variations of the calibration parameters between the datasets.

\section{Results and Discussion}
\label{sec:results}
%
%
In this section, we discuss the results of our experiments (\refsec{sec:experiments}) along the following questions:
\begin{itemize}
    \item \refsec{sec:result_rmse_sparsificaiton}: How accurate are motion estimates based on calibrations derived only from informative segments? How does it compare to the non-sparsified (batch) calibration?
    \item \refsec{sec:results_repeat}: Does the sparsified calibration yield similar calibration parameters to the (full) batch problem?
    \item \refsec{sec:result_accum_of_info}: Can we accumulate informative segments from multiple sessions and perform a calibration where the individual session would not provide enough excitation for a reliable calibration?
    \item \refsec{sec:results_comp_ekf}: How does the proposed method compare against an \ac{EKF} approach that jointly estimates motion and calibration parameters? 
    \item \refsec{sec:results_comp_metrics}: How do the three different information metrics compare? Can we outperform random selection of segments? 
    \item \refsec{sec:results_selected_segments}: What segments are being selected as informative? What are their properties?
    \item \refsec{sec:results_inf_dbsize}: How do we select the number of segments to retain in the database?
\end{itemize}

\begin{table*}[t]
\centering
\caption{
Comparison of the motion estimation accuracy evaluated on \ac{VIO} estimates when run with different calibration strategies.
The errors are shown for calibrations obtained from a sparsified problem ($8$ segments, $4$ seconds each) for three different information metrics and random selection as a baseline.
For reference, the errors are given for the batch calibration (no sparsification), the initial calibration and a related \ac{EKF}-based approach that uses the same \ac{VIO} estimator but jointly estimates the calibration, the motion and scene structure (similar to \cite{li2014high}).
All values show the median and standard deviation of the \ac{RMSE} over all evaluation datasets using the calibration under investigation.
}
\label{tab:rmse_offline}
\scriptsize 
\resizebox{2.0\columnwidth}{!}{%
\begin{tabular}{r|r@{}c@{}l|r@{}c@{}l|r@{}c@{}l|r@{}c@{}l|r@{}c@{}l|r@{}c@{}l|r@{}c@{}l}
\toprule
\multicolumn{1}{l}{}          & \multicolumn{21}{c}{\textbf{\ac{RMSE} on VIO trajectory vs. motion capture ground-truth (translation {[}cm{]} / rotation {[}deg{]})}}                                                                                                                                                                                                                                                                                                                                                                                       \\ \midrule
\multicolumn{1}{l}{\textbf{}} & \multicolumn{3}{c|}{\multirow{2}{*}{\textbf{\begin{tabular}[c]{@{}c@{}}initial\end{tabular}}}} & \multicolumn{3}{c|}{\multirow{2}{*}{\textbf{\begin{tabular}[c]{@{}c@{}}no sparsification\end{tabular}}}} & \multicolumn{12}{c|}{\textbf{sparsified (8 segments, each 4 seconds)}}                                                                                                        & \multicolumn{3}{c}{\multirow{2}{*}{\textbf{\begin{tabular}[c]{@{}c@{}}\end{tabular}}}} \\ \cmidrule(lr){8-19}
\multicolumn{1}{l}{\textbf{}} & \multicolumn{3}{c|}{\textbf{calibration}}                                                                                        & \multicolumn{3}{c|}{\textbf{(batch)}}                                                                                              & \multicolumn{3}{c|}{\textbf{E-optimality}}         & \multicolumn{3}{c|}{\textbf{D-optimality}}      & \multicolumn{3}{c|}{\textbf{A-optimality}}    & \multicolumn{3}{c|}{\textbf{random}}     & \multicolumn{3}{c}{\textbf{joint EKF}}                                                                               \\ \midrule
\textbf{AR/VR: office room}    & 13.07                         & $~\pm~$                         & 9.10 cm                          & \textbf{1.50}                           & $~\boldsymbol{\pm}~$                          & \textbf{0.89 cm}                           & 1.62 & $~\pm~$ & 0.60 cm  & 1.76 & $~\pm~$ & 0.59 cm  & 1.79 & $~\pm~$ & 0.62 cm  & 3.99 & $~\pm~$ & 2.49 cm  & 1.86                       & $~\pm~$                       & 1.17 cm                       \\
                               & 1.18                         & $~\pm~$                         & 0.60 deg                         & 0.47                           & $~\pm~$                          & 0.26 deg                          & \textbf{0.34} & $~\boldsymbol{\pm}~$ & \textbf{0.12 deg} & 0.37 & $~\pm~$ & 0.13 deg & 0.35 & $~\pm~$ & 0.13 deg & 0.64 & $~\pm~$ & 0.34 deg & 0.49                       & $~\pm~$                       & 0.27 deg                      \\
\textbf{AR/VR: class room}     & 13.09                         & $~\pm~$                         & 9.13 cm                          & 1.41                           & $~\pm~$                          & 1.07 cm                           & 1.79 & $~\pm~$ & 0.75 cm  & \textbf{1.28} & $~\boldsymbol{\pm}~$ & \textbf{0.54 cm}  & 1.42 & $~\pm~$ & 0.57 cm  & 5.45 & $~\pm~$ & 5.81 cm  & 2.44                       & $~\pm~$                       & 1.71 cm                       \\
                               & 1.17                         & $~\pm~$                         & 0.59 deg                         & 0.46                           & $~\pm~$                          & 0.25 deg                          & 0.35 & $~\pm~$ & 0.12 deg & \textbf{0.34} & $~\boldsymbol{\pm}~$ & \textbf{0.12 deg} & 0.35 & $~\pm~$ & 0.12 deg & 0.77 & $~\pm~$ & 0.54 deg & 0.52                       & $~\pm~$                       & 0.32 deg                      \\ \midrule
\textbf{NAV: parking garage}   & 13.09                         & $~\pm~$                         & 9.13 cm                          & 4.66                           & $~\pm~$                          & 34.73 cm                           & 1.65 & $~\pm~$ & 0.56 cm  & 2.14 & $~\pm~$ & 1.03 cm  & \textbf{1.59} & $~\boldsymbol{\pm}~$ & \textbf{0.59} cm  & 4.97 & $~\boldsymbol{\pm}~$ & 3.56 cm  & 3.04                       & $~\pm~$                       & 1.81 cm                       \\
                               & 1.17                         & $~\pm~$                         & 0.59 deg                         & 0.57                           & $~\pm~$                          & 0.62 deg                          & \textbf{0.31} & $~\boldsymbol{\pm}~$ & \textbf{0.11} deg & 0.38 & $~\pm~$ & 0.14 deg & \textbf{0.31} & $~\boldsymbol{\pm}~$ & \textbf{0.11 deg} & 0.73 & $~\pm~$ & 0.43 deg & 0.55                       & $~\pm~$                       & 0.29 deg                      \\
\textbf{NAV: office building}  & 13.13                         & $~\pm~$                         & 9.17 cm                          & 1.86                           & $~\pm~$                          & 1.17 cm                           & 1.68 & $~\pm~$ & 0.62 cm  & 1.39 & $~\pm~$ & 0.49 cm  & \textbf{1.26} & $~\boldsymbol{\pm}~$ & \textbf{0.45 cm}  & 2.32 & $~\pm~$ & 1.18 cm  & 2.56                       & $~\pm~$                       & 1.60 cm                       \\
                               & 1.16                         & $~\pm~$                         & 0.57 deg                         & 0.51                           & $~\pm~$                          & 0.27 deg                          & 0.41 & $~\pm~$ & 0.14 deg & \textbf{0.34} & $~\boldsymbol{\pm}~$ & \textbf{0.12 deg} & 0.35 & $~\pm~$ & 0.12 deg & 0.50 & $~\pm~$ & 0.27 deg & 0.60                       & $~\pm~$                       & 0.35 deg                      \\ \bottomrule
\end{tabular}

}%
\end{table*}

\subsection{Motion Estimation Performance using the Observability-aware Sparsification (Single-session Database)}
\label{sec:result_rmse_sparsificaiton}
In this experiment, we use a database of $8$ segments ($4$ seconds each) which leads to a reduction of the sessions size by around $75$\% in the \textit{AR/VR use-case} and $90$\% in the \textit{navigation use-case}.
To evaluate the observability-aware sparsification, we select the most informative segments for all sessions of a dataset independently.
A segment-based calibration is then run over the selected segments to obtain an updated set of calibration parameters for each session.
Finally, the \ac{VIO} motion estimation accuracy is evaluated for each calibration on all of the $15$ evaluation datasets as described in \refsec{sec:exp_vicon}.
The resulting statistics of the \ac{RMSE} are shown in \reftab{tab:rmse_offline} for each dataset.
The mean of rotation states corresponds to the rotation angle of the averaged quaternion as described in \cite{markley2007averaging} and the standard deviation is derived from rotation angles between the samples and the averaged quaternion.
For comparison, the same evaluations have been performed for the initial and batch calibration (no sparsification).

The calibrations obtained with the sparsified dataset yield very similar motion estimation performance when compared to full batch calibrations.
This indicates that the proposed method can indeed sparsify the calibration problem while retaining the relevant portion of the dataset and still provide a calibration with motion estimation performance close to the non-sparsified problem.
It is interesting to note, that the sparsification to a fixed number of segments keeps the calibration problem complexity bounded while the complexity of the batch problem is (potentially) unbounded when used on large datasets with redundant and non-informative sections.

\subsection{Repeatability of Estimated Calibration Parameter}
\label{sec:results_repeat}
As we have no ground-truth for the calibration parameters, we can only evaluate their repeatability across multiple calibrations of the same device.
The statistics over all calibration parameters obtained with all sessions of the \textit{class room} datasets are shown in \reftab{tab:params_repeatability}.
We used the same sparsification parameters as in \refsec{sec:result_rmse_sparsificaiton} ($8$ segments, each $4$ seconds).

The experiments show that the deviation between the full-batch and sparsified solution remain insignificant in mean and standard deviation even though $75$\% of the trajectory has been removed.
This is a good indication that the sparsified calibration problem is a good approximation to the complete problem.

\begin{table}[t]
\centering
\caption{
Mean and standard deviation of the estimated calibration parameters for the sparsified calibration problem (8 segments, 4 seconds), the batch solution and the final estimate of the joint \ac{EKF} run on the complete dataset.
The statistics have been derived from the calibrations obtained on all session of the \textit{AR/VR use-case} dataset.
The joint \ac{EKF} only estimates the \ac{IMU} intrinsics and the camera-\ac{IMU} relative pose, therefore no values are given for the camera intrinsics.
}
\label{tab:params_repeatability}
\resizebox{\columnwidth}{!}{%
   \begin{tabular}
      { r r@{}c@{}l r@{}c@{}l r@{}c@{}l }\toprule
      \shortstack{\bf{parameter}} &\multicolumn{3}{c }{\bf{proposed method}} & \multicolumn{3}{c }{\bf{batch}} & \multicolumn{3}{c }{\bf{joint EKF}} \\
      &\multicolumn{3}{c }{(sparsified)} & \multicolumn{3}{c }{(complete dataset)} & \multicolumn{3}{c }{(complete dataset)} \\
      \midrule
    $\mathbf{f}$ [px] & 255.79 & ~$\pm$~ & 0.60 & 256.30 & ~$\pm$~ & 0.22 & - \\ 
~ & 255.68 & ~$\pm$~ & 0.67 & 256.31 & ~$\pm$~ & 0.27 & - \\ 
$\mathbf{c}$ [px] & 313.63 & ~$\pm$~ & 0.67 & 313.19 & ~$\pm$~ & 0.63 & - \\ 
~ & 241.62 & ~$\pm$~ & 1.17 & 243.16 & ~$\pm$~ & 0.18 & - \\ 
$w$ [-] & 0.9203 & ~$\pm$~ & 0.0009 & 0.9208 & ~$\pm$~ & 0.0008 & - \\ 
$\mathbf{s}_g - 1$ [-] & -2.82e-03 & ~$\pm$~ & 1.32e-03 & -2.11e-03 & ~$\pm$~ & 2.27e-04 & 2.39e-03 & ~$\pm$~ & 2.06e-03 \\ 
~ & 4.33e-03 & ~$\pm$~ & 4.83e-03 & 4.02e-03 & ~$\pm$~ & 2.70e-04 & 7.71e-03 & ~$\pm$~ & 3.08e-03 \\ 
~ & -1.21e-03 & ~$\pm$~ & 5.18e-04 & -1.54e-03 & ~$\pm$~ & 4.18e-04 & 2.61e-03 & ~$\pm$~ & 3.90e-03 \\ 
$\mathbf{s}_a - 1$ [-] & -9.70e-03 & ~$\pm$~ & 1.50e-02 & -1.85e-02 & ~$\pm$~ & 3.07e-03 & -1.64e-02 & ~$\pm$~ & 6.54e-03 \\ 
~ & -1.16e-02 & ~$\pm$~ & 1.17e-02 & -1.65e-02 & ~$\pm$~ & 1.19e-03 & -1.24e-02 & ~$\pm$~ & 5.59e-03 \\ 
~ & -1.95e-02 & ~$\pm$~ & 7.38e-03 & -1.86e-02 & ~$\pm$~ & 1.48e-03 & -1.34e-02 & ~$\pm$~ & 2.43e-03 \\ 
$\mathbf{m}_g$ [-] & -3.22e-04 & ~$\pm$~ & 1.69e-03 & 7.36e-04 & ~$\pm$~ & 6.56e-04 & 1.03e-03 & ~$\pm$~ & 8.78e-04 \\ 
~ & 2.37e-03 & ~$\pm$~ & 1.95e-03 & 3.96e-04 & ~$\pm$~ & 2.30e-04 & -7.36e-04 & ~$\pm$~ & 1.32e-03 \\ 
~ & -6.78e-04 & ~$\pm$~ & 1.60e-03 & -4.95e-05 & ~$\pm$~ & 1.17e-03 & -9.82e-04 & ~$\pm$~ & 1.77e-03 \\ 
$\gamma(\mathbf{q}_{GA})$ [deg] & 1.897 & ~$\pm$~ & 0.428 & 1.504 & ~$\pm$~ & 0.010 & 1.368 & ~$\pm$~ & 0.150 \\ 
$\mathbf{m}_a$ [-] & 2.11e-02 & ~$\pm$~ & 1.11e-02 & 1.35e-02 & ~$\pm$~ & 1.54e-03 & 1.68e-02 & ~$\pm$~ & 5.05e-03 \\ 
~ & -3.68e-02 & ~$\pm$~ & 1.11e-02 & -2.78e-02 & ~$\pm$~ & 2.59e-03 & -2.76e-02 & ~$\pm$~ & 6.78e-03 \\ 
~ & -7.93e-03 & ~$\pm$~ & 9.30e-03 & -3.19e-03 & ~$\pm$~ & 1.21e-03 & -7.92e-04 & ~$\pm$~ & 2.99e-03 \\ 
$_{C}\mathbf{p}_{IC}$ [m] & 1.06e-03 & ~$\pm$~ & 4.01e-03 & 4.93e-03 & ~$\pm$~ & 2.33e-03 & 5.43e-03 & ~$\pm$~ & 3.68e-03 \\ 
~ & 4.62e-03 & ~$\pm$~ & 1.86e-02 & 7.05e-04 & ~$\pm$~ & 2.17e-03 & 4.09e-04 & ~$\pm$~ & 2.85e-03 \\ 
~ & -1.48e-02 & ~$\pm$~ & 1.12e-02 & -6.09e-03 & ~$\pm$~ & 4.08e-03 & -1.19e-02 & ~$\pm$~ & 6.77e-03 \\ 
$\gamma(\mathbf{q}_{IC})$ [deg] & 1.174 & ~$\pm$~ & 0.133 & 1.065 & ~$\pm$~ & 0.071 & 0.753 & ~$\pm$~ & 0.069 \\ 

      \bottomrule
    \end{tabular}

}%
\end{table}

\subsection{Comparison of Information Metrics}
\label{sec:results_comp_metrics}
In \refsec{sec:eval_segment_info}, we have proposed three different information metrics to compare trajectory segments for their information \wrt to the calibration parameters.
The same evaluation performed for the sparsification use-case (\refsec{sec:result_rmse_sparsificaiton}) has been repeated for each of the proposed metrics and, as a baseline, also for calibrations based on randomly selected segments.
The motion estimation errors based on these calibration is reported in \reftab{tab:rmse_offline}.

The motion estimation error is around $2$-$3$ times larger when randomly selecting the same amount of data indicating that the proposed metrics successfully identify informative segments for calibration.
It is important to note, that this comparison heavily depends on the ratio of informative / non-informative motion in the dataset and therefore this error might be larger when there is less excitation in a given dataset.
In general, all three metrics show comparable performance, however, the A-optimality criteria performed slightly better on the \textit{navigation} and the D-optimality on the \textit{AR/VR use-case}.

\subsection{Accumulation of Information over Time: Single- vs Multi-session Database}
\label{sec:result_accum_of_info}
\begin{figure}
 \centering
 \includegraphics[width=9cm]{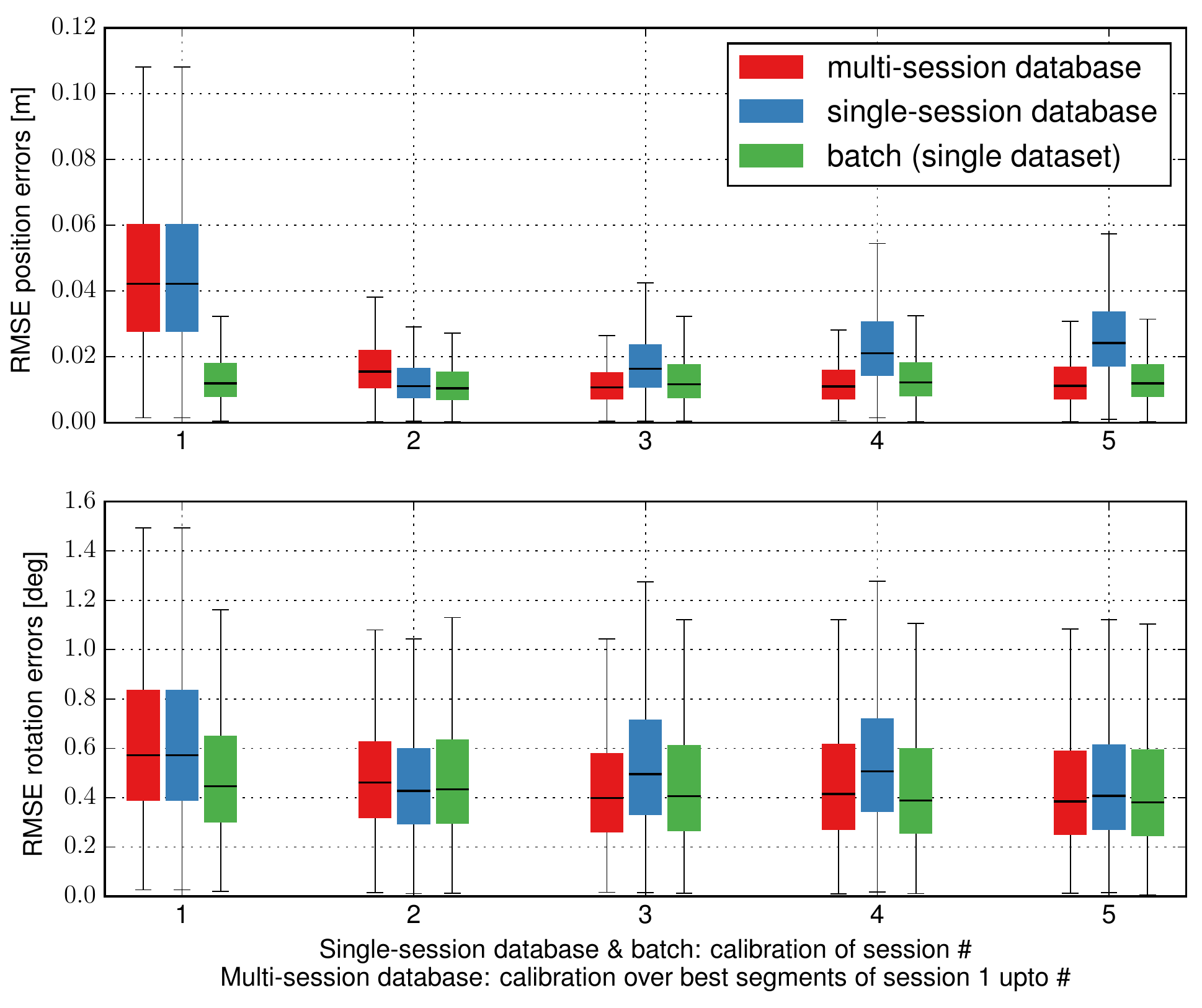}
 \caption{
 Comparing the \ac{VIO} motion estimation \ac{RMSE} for calibrations obtained with two different database strategies.
 A fixed number of the most informative segments ($8$ segments each $4$ seconds) have been collected either:
 (a) incrementally over all datasets (multi-session: \refsec{sec:llcal_mode_multi_session}), or
 (b) only from a single dataset (single-session: \refsec{sec:llcal_mode_single_session}).
 The motion estimation errors have been evaluated for all obtained calibrations based on these segments.
 For example, the calibration of session $3$ (x=3) and method (a), in red, is based on the $8$ most informative segments from the sessions $1$-$3$ and for method (b), in blue, on the $8$ most informative segments from session $3$ alone.
 The batch solution (green) uses all segments of a single dataset.
 }
 \label{fig:results_rmse_inc_vs_noninc}
\end{figure}

In this section, we evaluate whether the proposed method can accumulate informative segments from multiple consecutive sessions to obtain a better and more consistent calibration than the individual session would yield.
This is especially important in scenarios where a single session often would not provide enough excitation for a reliable calibration.
The evaluations were performed on the \textit{AR/VR use-case} datasets which consist of multiple short sessions.
We use the A-optimality criteria to select the most informative segments of each sessions and maintain them in the database ($8$ segments, $4$ seconds).
In contrast to the sparsification use-case from \refsec{sec:result_rmse_sparsificaiton}, the database is not reset between the sessions.
In other words, the database will collect the $N$ most informative segments from the first up to the current session.
After each session, a calibration is triggered using all segments of the database.
These calibrations are then used to evaluate the motion estimation error on all $15$ evaluation datasets.
The results are shown in \reffig{fig:results_rmse_inc_vs_noninc} for the \textit{class room} dataset.

The evaluation shows that the motion estimation error decreases as the number of sessions increases (from which informative segments have been selected).
Further, the motion estimation error is smaller when compared to calibrations based on the most informative segments from individual sessions. 
After around $2$ sessions the estimation performance is close to what would be achieved using a batch calibration.
This indicates that the proposed method can accumulate information from multiple sessions while the number of segments in the database remains constant.
It can therefore provide a reliable calibration when a single session would not provide enough excitation.

\subsection{Comparison vs. joint \ac{EKF}}
\label{sec:results_comp_ekf}
In this section, we compare the proposed method against an \ac{EKF} filter that jointly estimates the motion, scene structure and the calibration parameters (similar to \cite{li2014high}).
In our implementation, we only estimate the \ac{IMU} intrinsics and the relative pose between the camera and \ac{IMU}.
The camera intrinsics are not estimated and set to parameters obtained with a batch calibration on the same dataset.

\begin{figure}[t]
  \centering
  \includegraphics[width=0.48\textwidth]{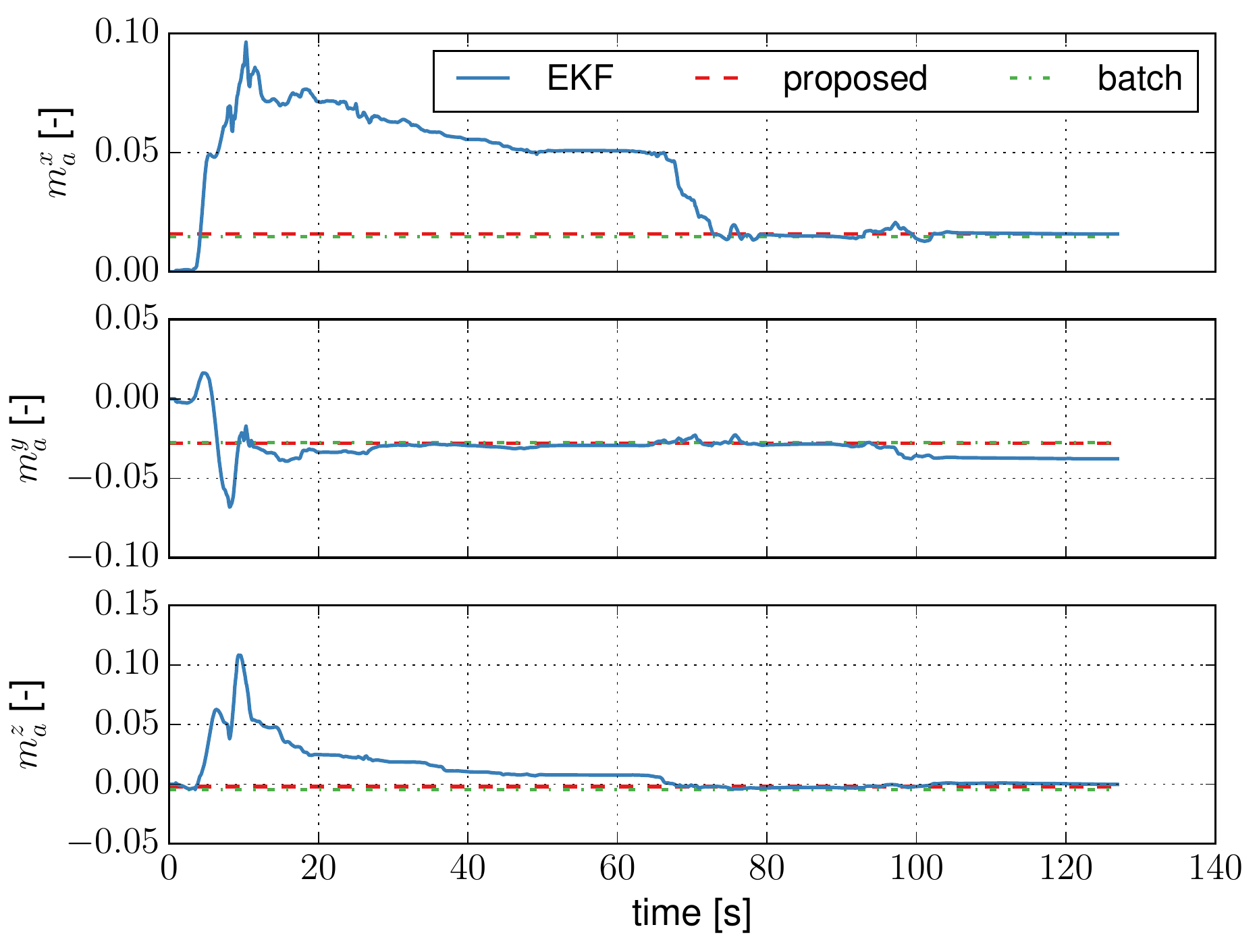}
  \caption{
      Misalignment of the gyroscope axis $\boldsymbol{m}_{g}$ estimated by the \ac{EKF} on one of the sessions in the \textit{class room} dataset.
      The \ac{EKF} jointly estimate the motion, structure and calibration parameters in a single filter.
      For comparison, the estimates obtained with the proposed method and the batch estimator are shown.
  }
  \label{fig:ekf_over_time}
\end{figure}

We evaluated the motion estimation errors on all datasets and report the results in \reftab{tab:rmse_offline}.
The resulting calibration parameters are compared to the proposed method and batch solution in \reftab{tab:params_repeatability}.
The evaluations show a position error that is up to $2$ times larger compared to calibrations obtained with the proposed method or a batch calibration.
When looking at the state evolution of e.g. the misalignment factors, as shown in \reffig{fig:ekf_over_time} for one of the datasets, it can be seen that it converges roughly to the batch estimate but does not remain stable over time.
We see this as an indication that the local scope of the \ac{EKF} is not able to infer weakly observable states properly and thus a segment-based (sliding-window) approach is beneficial in providing a stable and consistent solution over time.

\subsection{Selected Informative Segments}
\label{sec:results_selected_segments}
\begin{figure}[t]
\centering
 \includegraphics[width=0.5\textwidth]{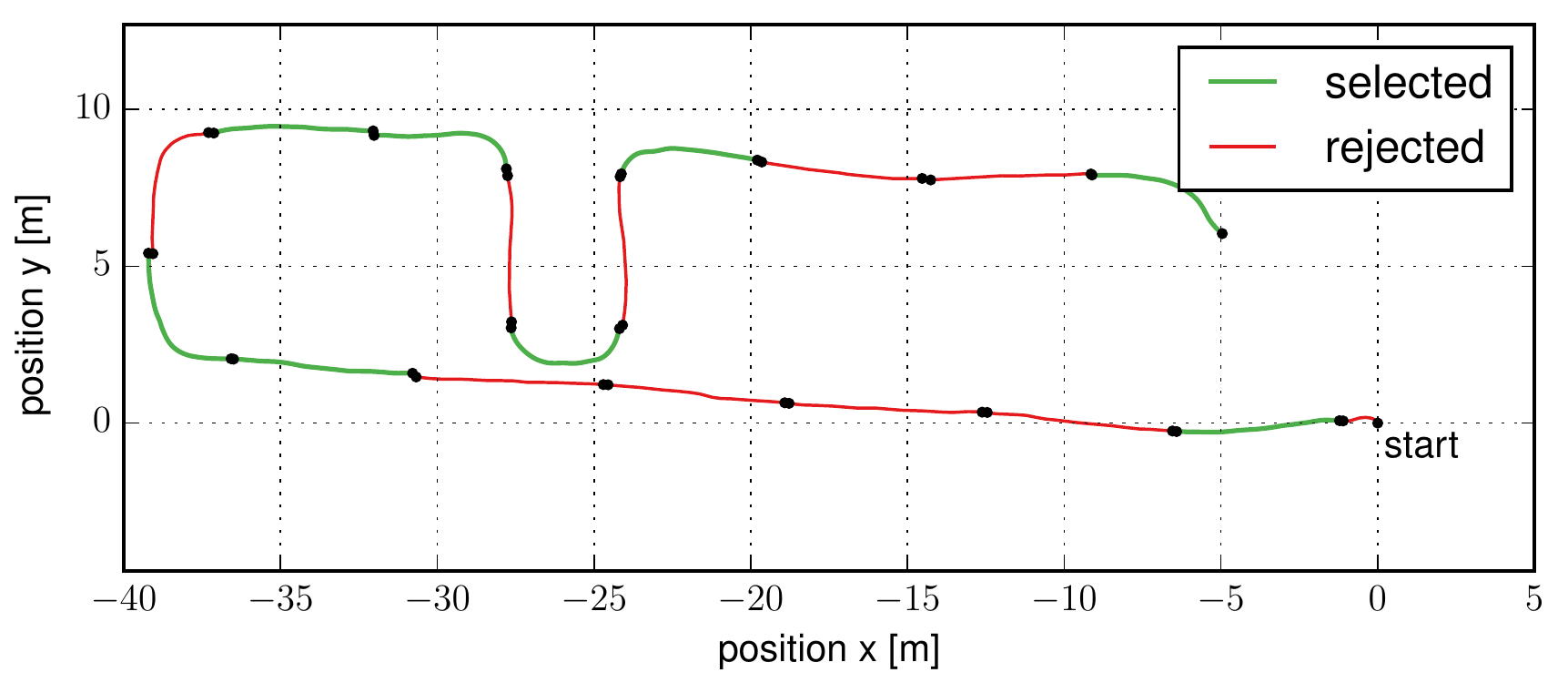}
\includegraphics[width=0.5\textwidth]{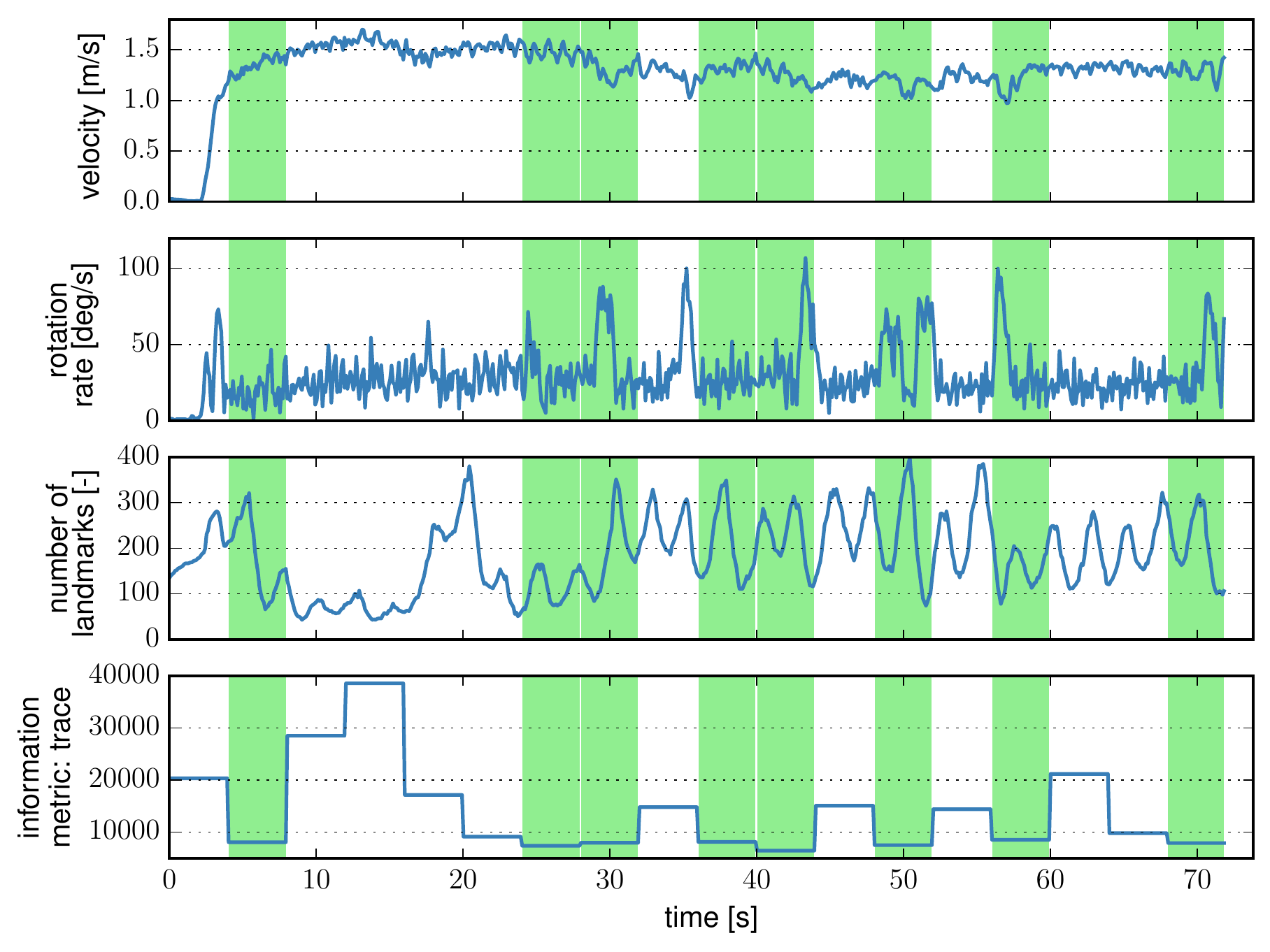}
\caption{%
The $8$ most informative segments identified using the A-optimality criteria in one of the sessions of the \textit{navigation use-case} (a lower value indicates more information).
The metric correlates with changes in the linear and rotational velocity and therefore mostly segments during turns have been selected whereas the straight segments were found to be less informative.
}
\label{fig:good_bad_selection}
\end{figure}

In this section, we investigate the motion that is being selected as informative by the proposed method.
\reffig{fig:good_bad_selection} shows the $8$ most informative segments that have been selected in one of the session of the \textit{navigation use-case}.
We only show the first minute of the session as otherwise the trajectory would start to overlap.
It can be seen that the information metric correlates with changes in linear and rotational velocity and therefore mostly segments containing turns have been selected while straight segments have been found to be less informative.
This experiment seems to confirm the intuition that segments with larger accelerations and rotational velocities are more informative for calibration.

\subsection{Influence of Database Size on the Calibration Quality}
\label{sec:results_inf_dbsize}
\begin{figure}[t]
\centering
\includegraphics[width=0.5\textwidth]{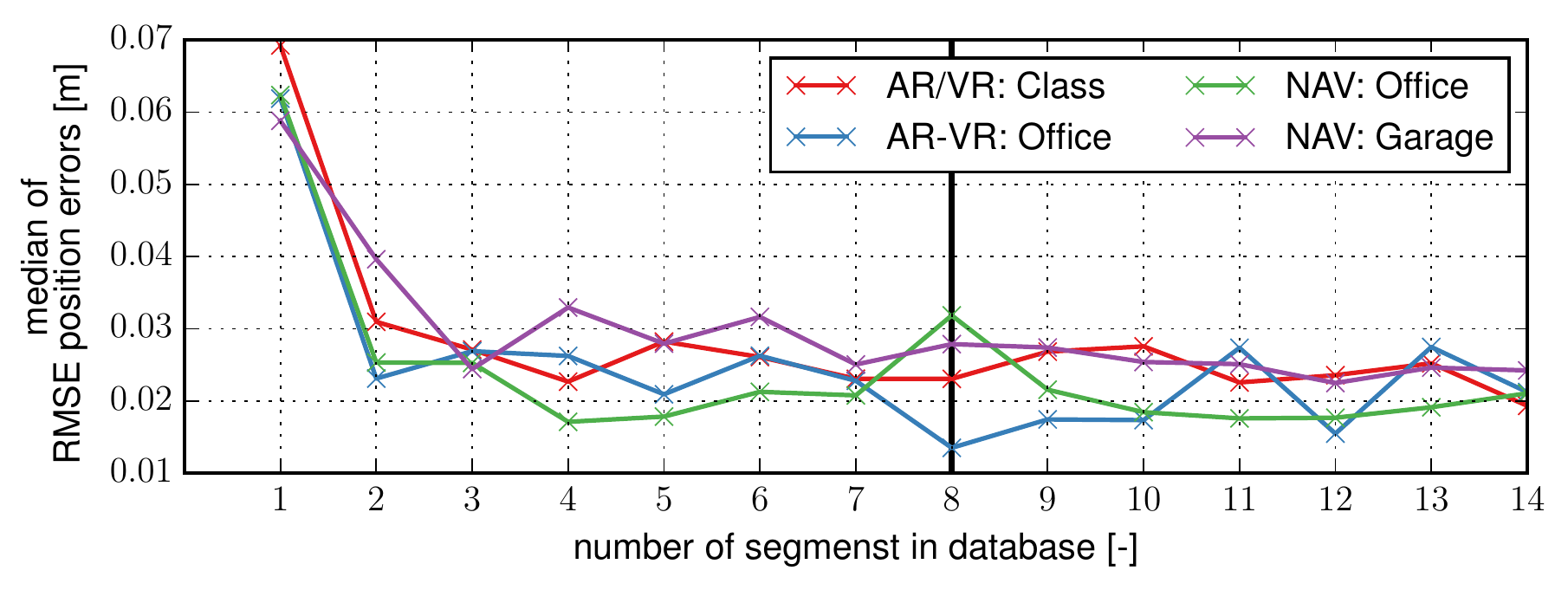}
\caption{%
Median of the motion estimation error for different levels of calibration datasets sparsification.
The error seems to stabilize when using more than $7-8$ segments and we found that $8$ segments provides a reasonable trade-off between complexity and quality.
}
\label{fig:rmse_over_db_size}
\end{figure}

In this experiment, we investigate the effect of the database size on the calibration quality to find the minimum amount of data required for a reliable calibration.
We sparsify all sessions of all datasets repeatably to retain $1$ to $15$ of the most informative segments.
A segment-based calibration is then run on each of the sparsified datasets and the motion estimation error is evaluated on all evaluation datasets.
The segment duration was chosen as $4$ seconds from geometrical considerations such that segments span a sufficiently large distance for landmark triangulation with the assumption that the system moves at a steady walking speed.
The median of the \ac{RMSE} over all evaluation datasets is shown in \reffig{fig:rmse_over_db_size}.

The motion estimation error seems to stabilize when using more than $7-8$ segments.
Based on these experiments, we have selected a database size of $8$ segments as a reasonable trade-off between calibration complexity and quality and used this value for all the evaluations in this work. 
It is important to note, that the amount of data required for a reliable calibration depends on the sensor models, the expected motion and the environment and a re-evaluation might become necessary if these parameters change.
In future work, we plan to investigate methods to determine the information content of the database directly to avoid a selection of this parameter.

\subsection{Run-time}
\reftab{tab:runtime} reports the measured run-times of the proposed method and the batch calibration for the experiments of \refsec{sec:result_rmse_sparsificaiton}.
Both optimizations use the same number of steps and the same initial conditions. 

It is important to note, that the complexity and thus run-time of the batch method is unbounded when the duration of the sessions increase.
The run-time of the proposed method, however, remains constant as we only include a constant amount of informative data.
This property makes the proposed method well-suited for systems performing long sessions.

\begin{table}[t]
\caption{%
Evaluation of the run-time for the proposed method, batch estimator and joint-\ac{EKF} obtained while running the experiments of \refsec{sec:result_rmse_sparsificaiton}.
The run-time of the batch calibration is unbounded as the calibration dataset increase.
The run-time of the proposed method, however, only depends on the number of collected informative segments and therefore has an upper bound.
}
\label{tab:runtime}
\centering
\begin{tabular}{@{}lccc@{}}
\toprule
 & \begin{tabular}[c]{@{}c@{}}\textbf{proposed}\\\textbf{method}\end{tabular} & \multicolumn{1}{l}{\textbf{batch}} & \begin{tabular}[c]{@{}c@{}}\textbf{joint}\\ \textbf{EKF} \end{tabular} \\ \midrule
\textbf{VIO} (each image)              & 0.003 s                                                     & -                         & 0.003 s                                                      \\
\textbf{Data selection} (each segment) & 0.156 s                                                    & -                         & -                                                           \\
\textbf{Calibration} (each dataset)    & 12.050 s                                                   & 27.028 s                    & -                                                           \\ \bottomrule
\end{tabular}
\end{table}

\section{Conclusion}
\label{sec:conclusion}
We have proposed an efficient self-calibration method for visual and inertial sensors which runs in parallel to an existing motion estimation framework.
In a background process, an information-theoretic metric is used to quantify the information content of motion segments and a fixed number of the most informative are maintained in a database.
Once enough data has been collected, a segment-based calibration is triggered to update the calibration parameters.
With this method, we are able to collect exciting motion in a background process and provide reliable calibration with the assumption that such motion occurs eventually - making this method well-suited for consumer devices where the users often do not know how to excite the system properly.

An evaluation on motion capture ground-truth shows that the calibrations obtained with the proposed method achieve comparable motion estimation performance to full batch calibrations.
However, we can limit the computational complexity by only considering the most informative part of a dataset and thus enable calibration even on long sessions and resource-constrained platforms where a full-batch calibration would be unfeasible.
Further, our evaluations show that we can not only sparsify single-session datasets but also accumulate information from multiple sessions and thus perform reliable calibrations when a single-session would not provide enough excitation.
The comparison of three information metrics indicates that A-optimality could be selected for navigation purposes while D-optimality looks like a good compromise for AR/VR applications.

In future work, we would like to investigate methods to dynamically determine the segment boundaries instead of using a fixed segment length and also account for temporal variations in the calibration parameters by detecting and removing outdated segments from a database.

\section*{Acknowledgements}
\label{sec:ack}
We would like to thank Konstantine Tsotsos, Michael Burri and Igor Gilitschenski for the valuable discussions and inputs. 
This work was partially funded by Google's Project Tango.

\bibliographystyle{IEEEtranN}
\bibliography{IEEEabrv,bibliography}

\vfill
\end{document}